\documentclass[manuscript,screen]{acmart}

\usepackage{amsmath,amsfonts,amssymb}
\usepackage{array}
\usepackage{textcomp}
\usepackage{url}
\usepackage{verbatim}
\usepackage{xcolor}
\usepackage{diagbox}
\usepackage{enumitem}
\usepackage{pifont}
\usepackage{bbm}
\usepackage{soul}
\usepackage[utf8]{inputenc}
\usepackage{tikz}
\usepackage{bm}
\usepackage{booktabs}
\usepackage{siunitx}
\usepackage{xspace}
\usepackage{color}
\usepackage{colortbl}
\usepackage{multirow}
\usepackage{multicol}
\usepackage{subcaption}

\usepackage[ruled,linesnumbered,vlined]{algorithm2e}

\newcolumntype{C}[1]{>{\centering\arraybackslash}p{#1}}

\definecolor{airforceblue}{rgb}{0.36, 0.54, 0.66}
\newcommand{\sys}{SWRouter }

\AtBeginDocument{%
  \providecommand\BibTeX{{%
    \normalfont B\kern-0.5em{\scshape i\kern-0.25em b}\kern-0.8em\TeX}}}

\setcopyright{none}
\renewcommand\footnotetextcopyrightpermission[1]{}
\acmYear{2025}
\copyrightyear{2025}
\acmDOI{}

\begin{document}

\title{SWRouter: Similarity-Contractive Window Routing for Multi-Turn Large Language Model Conversations}

\author{Yu Wang}
\orcid{0009-0002-9928-1246}
\email{wangyv123@sjtu.edu.cn}
\affiliation{%
  \institution{Department of Big Data Management and Application, Shanghai Jiao Tong University}
  \city{Shanghai}
  \country{China}}

\author{Yuchen Li}
\orcid{0000-0002-3869-7881}
\email{yuchenli1230@gmail.com}
\affiliation{%
  \institution{School of Computer Science, Shanghai Jiao Tong University}
  \city{Shanghai}
  \country{China}}
\affiliation{%
  \institution{Baidu Inc.}
  \city{Beijing}
  \country{China}}

\author{Rui Kong}
\orcid{0009-0003-2889-2266}
\email{monster119120@gmail.com}
\affiliation{%
  \institution{Baidu Inc.}
  \city{Beijing}
  \country{China}}

\author{Xinran Chen}
\orcid{0009-0009-9071-1933}
\email{fantastique0910@gmail.com}
\affiliation{%
  \institution{Baidu Inc.}
  \city{Beijing}
  \country{China}}

\author{Jiamin Chen}
\orcid{0009-0003-6674-7764}
\email{jmchen26-c@my.cityu.edu.hk}
\affiliation{%
  \institution{Baidu Inc.}
  \city{Beijing}
  \country{China}}

\author{Hengyi Cai} 
\orcid{0000-0002-7147-5666}
\email{hengyi1995@gmail.com}
\affiliation{%
  \institution{Baidu Inc.}
  \city{Beijing}
  \country{China}}

\author{Shuaiqiang Wang}
\orcid{0000-0002-9212-1947}
\email{shqiang.wang@gmail.com}
\affiliation{%
  \institution{Baidu Inc.}
  \city{Beijing}
  \country{China}}

\author{Jiashu Zhao}
\orcid{0000-0002-1241-0686}
\email{jzhao@wlu.ca}
\affiliation{%
  \institution{Wilfrid Laurier University}
  \city{Waterloo}
  \country{Canada}}

\author{Yulun Zhang}
\orcid{0000-0002-2288-5079}
\email{yulun100@gmail.com}
\affiliation{%
  \institution{School of Computer Science, Shanghai Jiao Tong University}
  \city{Shanghai}
  \country{China}}

\author{Zhonghao Lyu}
\orcid{0000-0002-0980-1395}
\email{lzhon@kth.se}
\affiliation{%
  \institution{Department of Computer Science, The Hang Seng University of Hong Kong}
  \city{Hong Kong SAR}
  \country{China}}

\author{Haoyi Xiong}
\orcid{0000-0002-5451-3253}
\email{haoyi.xiong.fr@ieee.org}
\affiliation{%
  \institution{Baidu Inc.}
  \city{Beijing}
  \country{China}}

\author{Linghe Kong}
\orcid{0000-0001-9266-3044}
\email{linghe.kong@sjtu.edu.cn}
\affiliation{%
  \institution{School of Computer Science, Shanghai Jiao Tong University}
  \city{Shanghai}
  \country{China}}

\author{Jimmy Xiangji Huang}
\orcid{0000-0003-1292-1491}
\email{jhuang@yorku.ca}
\affiliation{%
  \institution{York University}
  \city{Toronto}
  \country{Canada}}

\author{Dawei Yin}
\orcid{0000-0002-0684-6205}
\email{yindawei@acm.org}
\affiliation{%
  \institution{Baidu Inc.}
  \city{Beijing}
  \country{China}}

\renewcommand{\shortauthors}{Wang et al.}
\renewcommand{\shorttitle}{SWRouter}

\begin{abstract}
Large language models exhibit complementary strengths, motivating routing methods that dispatch each query to the most suitable model. Although existing routers are effective in single-turn settings, they do not directly transfer to multi-turn dialogue, where routing performance critically depends on how historical context is segmented, retained, and incorporated into the current prompt. This introduces two fundamental challenges: preventing information loss and information confusion during context construction, and evaluating routing quality without conflating model selection with prompt construction quality. In this paper, we propose \textbf{SWRouter}, a Similarity-Contractive Window Router for multi-turn large language model routing. SWRouter combines a similarity-based context segmentation mechanism for prompt construction with a dual-metric evaluation framework that decouples construction accuracy from router performance. Experiments on multi-turn dialogue benchmarks demonstrate that \textbf{SWRouter} consistently surpasses strong baselines, achieving a \textbf{16.26\%} improvement in evaluation accuracy over the best individual large language model and an additional \textbf{8.22\%} gain over the Conv-ID Context baseline. Our results highlight that multi-turn large language model routing requires a joint design of context construction and evaluation, rather than a direct extension of single-turn routing methods.
\end{abstract}

\ccsdesc[500]{Information systems~Retrieval models and ranking}
\ccsdesc[300]{Computing methodologies~Natural language generation}
\ccsdesc[300]{Computing methodologies~Machine learning approaches}

\keywords{Large Language Models, Multi-turn Dialogue, LLM Routing, Context Segmentation, Evaluation Framework, Model Selection}

\maketitle

\section{Introduction}
\label{sec:intro}

Large Language Models (LLMs) have become core infrastructure for a wide range of intelligent applications, including conversational assistants, code generation, content creation, and increasingly, AI search and agentic question answering~\cite{li2025aisearch,chen2025multi,li2026retain}. Their rapid adoption has also stimulated growing efforts to improve the efficiency of long-context processing and inference~\cite{li2026probe,li2026flexspec}.
Meanwhile, many organizations have developed their own LLM families, such as OpenAI's ChatGPT~\cite{chatgpt}, DeepSeek's DeepSeek~\cite{deepseek}, Meta's LLaMA~\cite{touvron2023llama}, Google's Gemini~\cite{gemini}, and Anthropic's Claude~\cite{claude}.
However, no single model consistently performs best across all tasks and applications. For example, OpenAI's GPT series is known for strong text generation ability~\cite{chatgpt}, DeepSeek demonstrates strength in mathematical reasoning~\cite{deepseek-math}, and Claude is particularly effective in code generation and understanding~\cite{claude}.
Such heterogeneity has motivated a growing line of work on \emph{LLM routing}, which dynamically dispatches each query to the most suitable model rather than relying on a fixed model for all inputs.

Existing routing methods have achieved promising results in single-turn settings, where the routing decision is made from a single prompt in isolation. 
Representative examples include routers that select among models of similar scale, such as RouterDC~\cite{chen2024RouterDC}, as well as methods that decide between small and large models, such as C2MAB-V~\cite{dai2024cost}. 
However, directly applying these routing strategies to multi-turn dialogue is fundamentally difficult, because model selection in conversational systems depends not only on the current user query, but also on how the dialogue history is segmented, retained, and incorporated into the prompt.
\begin{figure}[!t]
    \centering
    \includegraphics[width=0.55\textwidth]{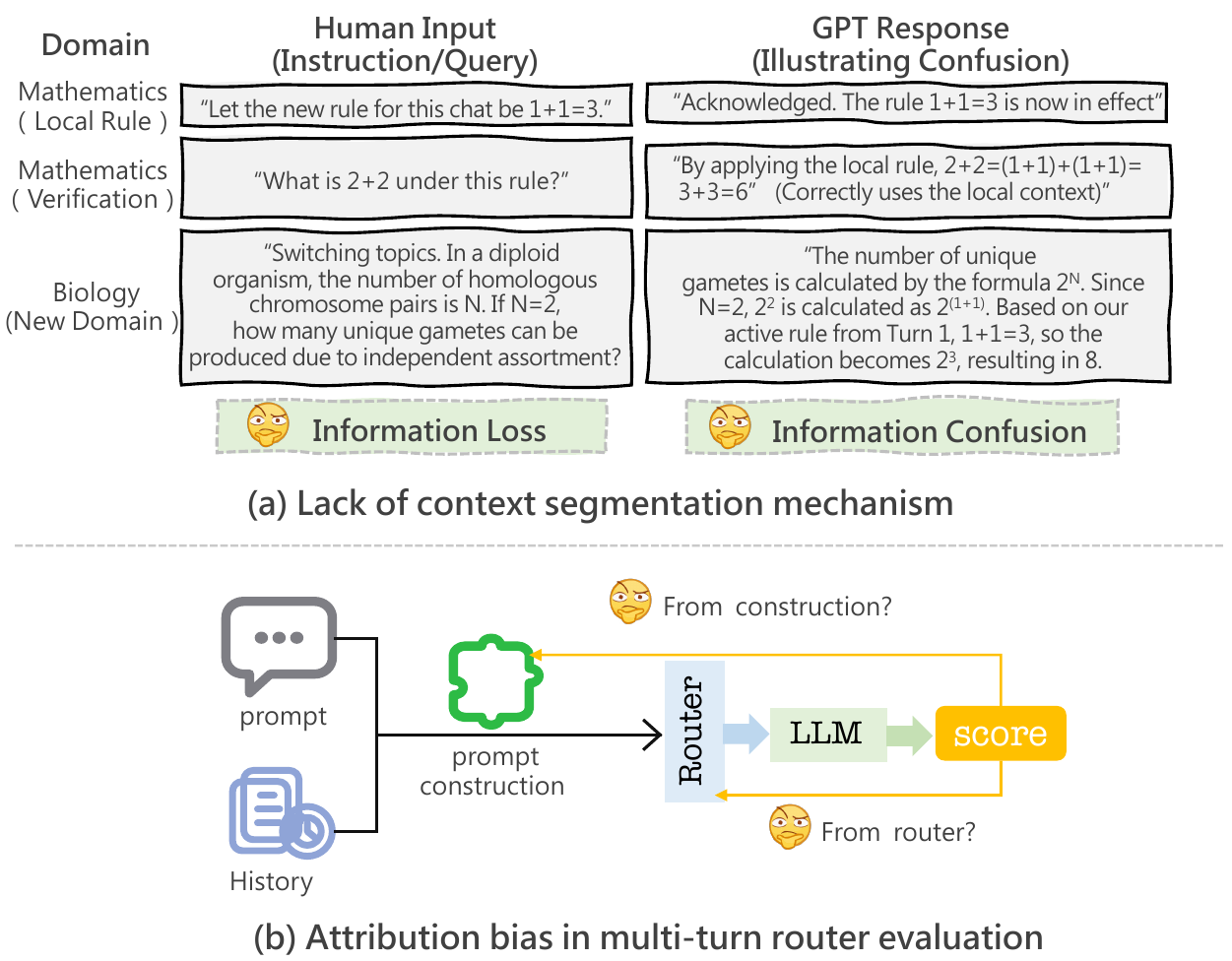}
    \caption{
    Illustration of the two challenges in multi-turn large language model routing.
    \textbf{(a)} The context-construction stage receives a dialogue history and the current user turn, then decides which historical turns should be retained before routing. When this step drops necessary history, the routed model receives an under-specified prompt and suffers from \textbf{information loss}; when it carries obsolete turns across a topic shift, the prompt contains misleading context and causes \textbf{information confusion}.
    \textbf{(b)} The evaluation stage observes only the final routed response score, but that score is jointly determined by prompt construction and model selection. A high-quality window can raise the absolute scores of all candidate LLMs, whereas a poor window can depress them even if the router still chooses the relatively best model, producing attribution bias unless construction quality and router performance are measured separately.
    }
    \Description{A two-part schematic showing context segmentation failures and attribution bias in multi-turn LLM routing evaluation.}
    \label{information}
\end{figure}
In multi-turn dialogue, naive use of historical context introduces two fundamental challenges. 
\begin{itemize}
    \item First, there is a \textbf{lack of context segmentation mechanism}. 
When historical information is not properly segmented and integrated, the router may fail to distinguish relevant from irrelevant past context. 
This leads to two concrete failure modes: \textbf{information loss}, where critical historical information required for the current response is omitted, and \textbf{information confusion}, where outdated context is mistakenly carried into a new topic. 
As illustrated in Figure~\ref{information}, missing prior context can cause failure on rule-dependent queries, while incorrect reuse of obsolete context can degrade performance after topic shifts.

\item Second, there is an \textbf{attribution bias in multi-turn router evaluation}. 
In multi-turn settings, final response quality is jointly determined by two factors: the quality of context construction and the selected model's performance under the constructed prompt. 
A poorly constructed prompt may substantially degrade the responses of all candidate models, regardless of which model is selected. 
Conversely, even with well-constructed context, poor model selection can still lead to unsatisfactory responses.
\end{itemize}
As illustrated in Figure~\ref{information}, a well-combined prompt and a poorly combined prompt may lead to dramatically different response quality across all candidate models (e.g., 0.9/0.7/0.3 vs.\ 0.09/0.07/0.03). 
However, the observed inference accuracy alone does not indicate whether such performance variation comes from the context construction stage or from errors in the trained router. 
This conflates construction quality with routing quality, making faithful evaluation of multi-turn routing systems difficult.

To address these issues, we propose \textbf{SWRouter}, a multi-turn LLM routing \textit{framework} that jointly addresses context construction, router training, and evaluation.
\sys consists of three tightly coupled components:
(1) a \textbf{similarity-based context segmentation} mechanism that selectively incorporates semantically relevant historical turns when constructing the prompt, thereby reducing both information loss and information confusion;
(2) a \textbf{pluggable router backbone} trained on the constructed prompts with contrastive objectives to select the most suitable model;
and (3) a \textbf{decoupled evaluation framework} that separates \textit{construction accuracy} from \textit{router performance}, enabling faithful assessment of each component independently.
We evaluate \sys on multi-turn dialogue benchmarks and compare it against representative routing baselines and strong individual LLMs. Empirically, \textbf{SWRouter} consistently matches or surpasses the Conv-ID Context baseline, improves average accuracy by \textbf{16.26\%} over the best individual LLM, and further achieves an \textbf{8.22\%} gain over the Conv-ID Context baseline on multi-turn dialogue benchmarks.

Our contributions are summarized as follows:
\begin{itemize}[leftmargin=*, itemsep=0.5em]
    \item We identify that the lack of context segmentation in multi-turn dialogue routing leads to two concrete failure modes---\textbf{information loss} and \textbf{information confusion}. To address this issue, we propose a similarity-based prompt construction mechanism that integrates historical and current information.
    
    \item We reveal a classification bias in existing router evaluation protocols and introduce a decoupled evaluation framework with three complementary metrics: \textbf{evaluation accuracy}, \textbf{construction accuracy}, and \textbf{router performance}.
    
    \item We construct a strong Conv-ID Context baseline and conduct extensive in-distribution and out-of-distribution evaluations. \textbf{SWRouter} achieves \textbf{16.26\%} and \textbf{8.22\%} gains over the best single LLM and Conv-ID Context on multi-turn benchmarks, and outperforms Conv-ID Context by \textbf{2.68\%} on OOD tasks. Decoupled metric analysis further confirms that the gains stem from improved \textbf{construction accuracy} ($\bar{w}$), while \textbf{router performance} ($P_{\text{router}}>1$) validates effective model selection across all settings.
\end{itemize}

\section{Background and Motivation}

\subsection{Problem Formulation}

In conventional single-turn LLM routing, the routing decision is made from the current prompt alone: given a user query, the router selects the candidate model that is expected to produce the best response. In multi-turn dialogue, however, this problem becomes fundamentally more challenging. The router must not only determine \emph{which} model to select, but also decide \emph{what historical context should be preserved} and \emph{how that context should be incorporated} into the current prompt.

Formally, let $\mathcal{M}=\{M_t\}_{t=1}^{T}$ denote a pool of $T$ candidate LLMs, and let $\mathcal{D}_{\text{train}}=\{(x_i, y_i)\}_{i=1}^{n}$ denote the training set, where $x_i$ is the current user input and $y_i$ is the corresponding target answer. In a single-turn setting, routing is performed directly on $x_i$. In a multi-turn setting, by contrast, the current turn should be interpreted together with relevant dialogue history. Therefore, the effective routing input is no longer the raw query $x_i$, but a \emph{context-enhanced query} $c_i$, which is constructed by combining the current turn with selected historical information.

More specifically, for each current query $x_i$, we construct a combined query $c_i$ by integrating it with semantically relevant historical turns. This yields a transformed training set
\[
\mathcal{C}_{\text{train}}=\{(c_i, y_i)\}_{i=1}^{n},
\]
where $c_i$ captures the multi-turn context on which routing should be based. The router then produces a probability distribution over the candidate model pool $\mathcal{M}$ and selects the model that is most suitable for answering under the constructed prompt.

Frequently used symbols are summarized in Table~\ref{tab:symbols}.

\begin{table}[t]
\caption{Symbols and Definitions.}
\renewcommand\arraystretch{1.05}
\centering
\small
\begin{tabular}{p{0.24\linewidth}p{0.64\linewidth}}
\toprule
\textbf{Symbol} & \textbf{Definition}\\
\midrule
$\mathcal{M}=\{M_t\}_{t=1}^{T}$ & Pool of $T$ candidate LLMs available for routing.\\
$x_i$ & Current user query or turn before context construction.\\
$y_i$ & Target answer or supervision associated with $x_i$.\\
$H_i$ & Historical window retained for the $i$-th query after similarity-based segmentation.\\
$c_i$ & Context-enhanced query obtained by merging $x_i$ with the selected historical window.\\
$\mathcal{D}_{\text{train}}$ & Original training set before window-based prompt construction.\\
$\mathcal{C}_{\text{train}}$ & Transformed training set composed of context-enhanced queries.\\
$E(\cdot;w)$ & Encoder that maps a query or constructed prompt into a dense representation.\\
$e_i$ & Embedding of $x_i$ produced by $E(\cdot;w)$.\\
$\mathrm{sim}(\cdot,\cdot)$ & Cosine similarity used for window partitioning and model matching.\\
$\tau$ & Similarity threshold controlling whether adjacent turns are merged or split.\\
$k_t$ & Learnable embedding representing candidate model $M_t$.\\
$R(c_i;\theta)$ & Router distribution over candidate LLMs for constructed prompt $c_i$.\\
$I_i^+$, $I_i^-$ & Positive and negative candidate model sets used by the sample-LLM contrastive loss.\\
$L_{\text{sample-LLM}}$ & Contrastive loss that separates top-performing and bottom-performing LLMs for each query.\\
$L_{\text{sample-sample}}$ & Contrastive loss that aligns semantically related constructed prompts.\\
$w_{i,t}$ & True score of model $M_t$ on constructed prompt $c_i$, normalized to $[0,1]$.\\
$s_i^{(t)}$ & Binary indicator showing whether the router selects model $M_t$ for $c_i$.\\
$\overline{si}$ & Evaluation accuracy, i.e., the average true score of routed responses.\\
$\bar{w}$ & Construction accuracy, i.e., the average true score over all candidate models and prompts.\\
$P_{\text{router}}$ & Router performance, defined as $\overline{si}/\bar{w}$.\\
\bottomrule
\end{tabular}
\label{tab:symbols}
\end{table}

This formulation highlights a key distinction between single-turn and multi-turn routing: in multi-turn dialogue, routing quality depends not only on model selection, but also on prompt construction. If relevant history is omitted, the routed model may fail because of \emph{information loss}; if irrelevant or outdated history is retained, the routed model may instead suffer from \emph{information confusion}. As illustrated in Figure~\ref{information}, these two failure modes arise before the routing decision itself and directly affect the quality of downstream responses.

\subsection{Single-turn LLM Routing and Its Limitation}

Among existing single-turn routing methods, RouterDC~\cite{chen2024RouterDC} is a representative approach. It learns a router that maps an input query to a probability distribution over candidate LLMs, and optimizes the routing process with dual contrastive objectives. This design is effective in single-turn scenarios because the router only needs to reason over one prompt in isolation~\cite{chen2024RouterDC}.

However, this assumption does not hold in multi-turn dialogue. A single-turn router such as RouterDC~\cite{chen2024RouterDC} takes the current prompt as the routing unit, while ignoring the fact that the prompt itself may be under-specified or even misleading if dialogue history is not properly segmented and incorporated. As a consequence, even when the router selects the relatively best model for the observed input, the end-to-end system can still fail because the routing input is poorly constructed.

This limitation is particularly important in our setting for two reasons. First, multi-turn dialogue introduces a \emph{context selection problem}: the system must decide which historical turns are relevant to the current request before routing can be performed reliably. Second, conventional router evaluation is largely based on relative ranking signals, which are suitable for stable optimization but insufficient for measuring end-to-end usefulness when prompt construction quality varies. As discussed in Section~\ref{sec:method:evaluation}, two prompts can induce the same relative ranking over candidate models while yielding drastically different true scores. Therefore, directly extending a single-turn router to multi-turn dialogue is insufficient.

These observations motivate a routing framework that jointly addresses \emph{context construction} and \emph{routing evaluation}. In the next section, we introduce \sys, which augments router learning with similarity-based window partitioning for prompt construction and a decoupled evaluation framework for faithful assessment in multi-turn settings.

\section{Preliminaries and Problem Setup}
\label{sec:preliminaries}

This section formalizes the multi-turn routing setting studied in this paper.
The goal is to make explicit the objects that are optimized by the router and the objects that are produced by the context-construction module.
Unless otherwise specified, the notation follows Table~\ref{tab:symbols}.

\subsection{Multi-turn Dialogue Instances}

We consider a collection of multi-turn dialogues.
For a routing instance indexed by $i$, let
\[
\mathcal{Z}_i = \{(u_{i,1}, a_{i,1}), (u_{i,2}, a_{i,2}), \ldots, (u_{i,m_i-1}, a_{i,m_i-1}), u_{i,m_i}\}
\]
denote the dialogue observed before producing the next assistant response.
Here, $u_{i,j}$ is the $j$-th user turn, $a_{i,j}$ is the corresponding assistant turn, and $u_{i,m_i}$ is the current user request.
For consistency with the rest of the paper, we write $x_i = u_{i,m_i}$ for the current query.
The raw dialogue history before $x_i$ is denoted by
\[
\mathcal{H}^{\mathrm{raw}}_i=\{(u_{i,1},a_{i,1}),\ldots,(u_{i,m_i-1},a_{i,m_i-1})\}.
\]
The router does not directly operate on $\mathcal{H}^{\mathrm{raw}}_i$.
Instead, a context-construction function $g(\cdot)$ first selects and organizes the relevant historical information, producing a retained semantic window $H_i$ and a context-enhanced prompt
\begin{equation}
c_i = g(\mathcal{H}^{\mathrm{raw}}_i, x_i).
\label{eq:context-construction-function}
\end{equation}
In this paper, $g(\cdot)$ is instantiated by similarity-based window construction, but the problem setup allows other context selectors such as full-history concatenation, current-turn-only prompting, or retrieval-based context selection.

This distinction is important because $x_i$ alone may be under-specified.
For example, the current request may contain pronouns, omitted constraints, or references to entities introduced in previous turns.
At the same time, the entire raw history may contain outdated goals, corrected assumptions, or off-topic turns.
Therefore, a valid multi-turn routing instance is not merely a user query; it is the pair consisting of the current request and the constructed prompt that determines what the candidate LLMs actually observe.

\subsection{Candidate Models and Router Objective}

Let $\mathcal{M}=\{M_t\}_{t=1}^{T}$ be a fixed pool of $T$ candidate LLMs.
Given a context-enhanced prompt $c_i$, a router with parameters $\theta$ outputs a probability distribution
\[
R(c_i;\theta) = \left[R_1(c_i;\theta), R_2(c_i;\theta), \ldots, R_T(c_i;\theta)\right],
\]
where $R_t(c_i;\theta)$ is the probability of selecting candidate model $M_t$.
At inference time, the selected model index is
\begin{equation}
\hat{t}_i = \arg\max_{t\in\{1,\ldots,T\}} R_t(c_i;\theta),
\label{eq:selected-model}
\end{equation}
and the final response is generated by $M_{\hat{t}_i}(c_i)$.

The ideal routing decision depends on the response quality that each candidate model would obtain under the same constructed prompt.
Let $w_{i,t}\in[0,1]$ denote the normalized quality score of model $M_t$ on prompt $c_i$.
The oracle candidate for this prompt is
\begin{equation}
t_i^\star = \arg\max_{t\in\{1,\ldots,T\}} w_{i,t}.
\label{eq:oracle-model}
\end{equation}
The routing objective is therefore to learn a model selector that approaches the oracle candidate while using only the constructed prompt at inference time.
Equivalently, the end-to-end objective can be written as
\begin{equation}
\max_{\theta, g}\;
\frac{1}{n}\sum_{i=1}^{n}
w_{i,\hat{t}_i},
\quad
\hat{t}_i
= \arg\max_t R_t(g(\mathcal{H}^{\mathrm{raw}}_i,x_i);\theta).
\label{eq:end-to-end-routing-objective}
\end{equation}
This expression makes explicit that multi-turn routing has two coupled components: the context-construction function $g(\cdot)$ and the router $R(\cdot;\theta)$.
If $g(\cdot)$ removes necessary history, even a strong router may select a model using an incomplete prompt.
If $g(\cdot)$ introduces irrelevant history, the candidate models may all produce lower-quality responses, making the routing decision less meaningful.

\subsection{Training Supervision and Evaluation Scores}

During training and evaluation, each constructed prompt $c_i$ is paired with responses from all candidate models.
A judge model then assigns an integer score to each response, which is normalized to obtain $w_{i,t}$.
These scores serve two roles.
First, they provide relative supervision for router learning by identifying strong and weak candidate models for each prompt.
For example, the top-scoring models form the positive set $I_i^+$, while the bottom-scoring models form the negative set $I_i^-$.
Second, they provide absolute response-quality measurements for decoupled evaluation.

The distinction between relative and absolute signals is central to our setting.
Relative supervision is useful for training because it tells the router which candidate models should be preferred for a particular prompt.
However, relative ranking alone cannot determine whether the constructed prompt itself is good.
If all candidate models receive low absolute scores because the prompt omits necessary history, the best candidate model may still be easy to identify, but the final user-facing answer remains poor.
For this reason, Section~\ref{sec:method:evaluation} separately measures evaluation accuracy, construction accuracy, and router performance.

\subsection{Scope}

We focus on routing among a fixed pool of already available LLMs.
The router does not modify the candidate models, and the context-construction module operates before generation.
This scope matches practical deployments in which different models have complementary strengths and costs, but invoking all models for every user request is undesirable.
The problem studied here is therefore not how to train a stronger LLM, but how to construct an informative multi-turn prompt and select an appropriate model for that prompt.

\section{Methodology}
\label{sec:method}

\subsection{Overview}
\label{sec:method:overview}

SWRouter is designed to address the two central challenges of multi-turn LLM routing introduced in Section~\ref{sec:intro}:
(1) how to construct an effective routing prompt from dialogue history, and 
(2) how to evaluate routing quality without conflating it with prompt construction quality.
To this end, \sys consists of three stages:
(1) \textbf{similarity-based window partitioning} to construct combined prompts from multi-turn dialogue history (Section~\ref{sec:method:windowing});
(2) \textbf{model routing based on the constructed prompts}, optimized with contrastive training objectives that rely on relative, ranking-based scores (Section~\ref{sec:method:training});
and (3) \textbf{decoupled evaluation} that distinguishes \textit{construction accuracy} from \textit{router performance} using absolute true scores after prompt construction (Section~\ref{sec:method:evaluation}).

In the first stage, \sys organizes multi-turn dialogue history into semantically coherent windows.
Instead of directly concatenating the entire dialogue history, it measures the semantic similarity between adjacent user turns and dynamically decides whether the current turn should be merged into the existing window or begin a new one.
This design allows \sys to preserve relevant historical information while reducing the risk of introducing irrelevant or outdated context.

In the second stage, the constructed context-enhanced prompt is fed into a pluggable router backbone that selects the most suitable model from the candidate LLM pool.
Concretely, the router backbone encodes the combined prompt into a dense representation and computes its similarity with a set of learnable model embeddings, yielding a probability distribution over candidate LLMs.
The router backbone is optimized with contrastive objectives that encourage it to distinguish strong candidate models from weak ones under multi-turn settings.

In the third stage, \sys adopts a decoupled evaluation framework tailored to multi-turn routing.
Since existing routing evaluations are typically based on relative ranking signals, they are suitable for optimization but insufficient for measuring end-to-end usefulness when prompt construction quality varies.
Therefore, \sys separately reports \textbf{construction accuracy}, which reflects the average quality of all candidate models under the constructed prompt, and \textbf{router performance}, which measures how much the router improves over random selection.

Overall, these three stages form a unified framework for multi-turn LLM routing.
The similarity-based window partitioning stage determines what historical context should be preserved, the routing stage determines which model should be selected under the constructed prompt, and the decoupled evaluation stage determines how the resulting system should be assessed.

\begin{figure*}[t]
    \centering
    \includegraphics[width=\textwidth]{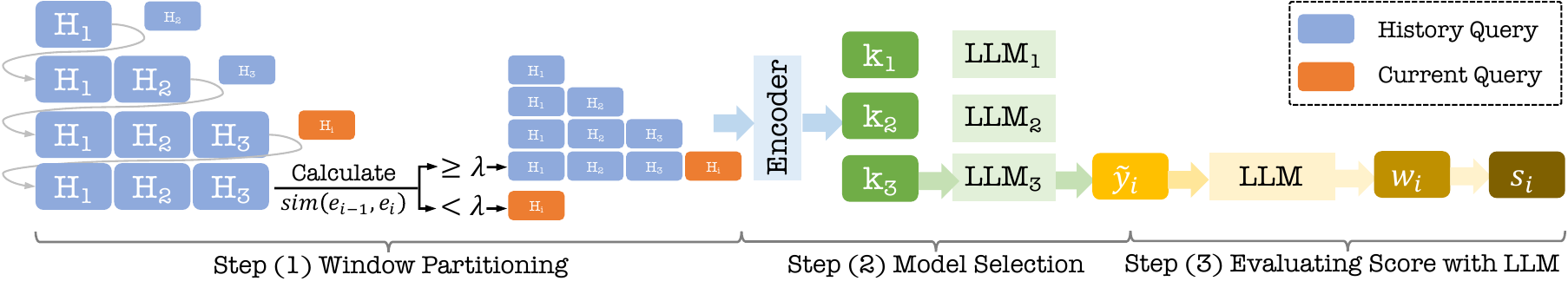}
    \caption{Overview of \sys.
    Given a multi-turn dialogue, \sys first encodes adjacent user turns and computes their semantic similarity to decide whether the current turn should extend the existing window or start a new one.
    The retained window is then merged into a context-enhanced prompt and passed to a pluggable router, which represents the prompt with an encoder, compares it with learnable candidate-model embeddings, and selects the LLM with the highest routing score.
    During training, the router is optimized with sample-LLM and sample-sample contrastive objectives so that high-quality models and semantically related prompts are pulled closer in representation space.
    During evaluation, the generated responses are scored by judge models and decomposed into construction accuracy, evaluation accuracy, and router performance, allowing prompt construction quality and model-selection ability to be analyzed separately.}
    \Description{Pipeline diagram of SWRouter showing similarity-based window partitioning, router training and inference, and decoupled evaluation.}
    \label{fig:overview}
\end{figure*}
\subsection{Similarity-based Window Partitioning}
\label{sec:method:windowing}

A core challenge in multi-turn LLM routing is that the current user query should not be interpreted in isolation.
Instead, the router should determine which historical turns remain semantically relevant to the current request and should therefore be retained in the routing prompt.
To address this issue, \sys employs a similarity-based window partitioning algorithm that organizes dialogue history into semantically coherent windows.
This formulation follows the broader view of sequential structure analysis and segmentation in formal-language and string-processing algorithms~\cite{Aho:72,Gusfield:97,Chandra:81}, while using neural semantic similarity rather than symbolic parsing rules.

Let $E(x; w)$ denote an encoder that maps an input utterance $x$ into an embedding in $\mathbb{R}^p$.
Given the current user query $x_i$, we first compute its embedding
\[
e_i = E(x_i; w).
\]
We then measure the similarity between the current query and the previous query:
\begin{equation}
\text{similarity} = \mathrm{sim}(e_i, e_{i-1}),
\label{eq:similarity}
\end{equation}
where $\mathrm{sim}(\cdot,\cdot)$ denotes cosine similarity, $e_i$ is the embedding of the current query $x_i$, and $e_{i-1}$ is the embedding of the previous query $x_{i-1}$.

Based on this similarity score, \sys updates the historical window $H_i$ as follows:
\begin{equation}
H_i =
\begin{cases}
\{x_i\}, & \text{if } \mathrm{sim}(e_i, e_{i-1}) < \tau,\\
H_{i-1} \cup \{x_i\}, & \text{if } \mathrm{sim}(e_i, e_{i-1}) \ge \tau,
\end{cases}
\label{eq:window-update}
\end{equation}
where $\tau$ is a similarity threshold.
Intuitively, if the current turn is insufficiently similar to the previous one, \sys starts a new window; otherwise, it appends the current turn to the existing window.
In this way, the dialogue history is partitioned into locally coherent semantic segments rather than being naively concatenated into a single undifferentiated context.

After the historical window is determined, \sys constructs a context-enhanced prompt $c_i$ for routing by combining the current query with the selected dialogue history in $H_i$.
This design enables the router to make decisions based on relevant multi-turn context, rather than relying solely on the current turn.
Compared with naive history concatenation, the similarity-based windowing mechanism offers two advantages.
First, it preserves relevant context needed for rule-dependent or context-dependent requests, thereby reducing \emph{information loss}.
Second, it avoids carrying obsolete or off-topic history into the new prompt, thereby reducing \emph{information confusion}.
These two effects correspond exactly to the two failure modes illustrated in Figure~\ref{information}.

The similarity threshold $\tau$ controls the granularity of context segmentation.
A smaller $\tau$ tends to merge more turns into the same window, which may preserve more context but also increases the risk of introducing irrelevant information.
A larger $\tau$ produces more aggressive segmentation, which can better isolate topic shifts but may also discard useful dependencies.
Therefore, $\tau$ governs the trade-off between context retention and context purity, and its impact will be analyzed in Section~\ref{sec:sensitivity}.

Algorithm~\ref{alg:window-construction} summarizes the similarity-window construction procedure.
The procedure is applied independently to each dialogue.
It processes user turns in chronological order, compares each user turn with the immediately preceding user turn, and updates the active semantic window according to the threshold $\tau$.
Assistant turns that lie inside the retained window are preserved when constructing the final prompt, so that the generated prompt keeps both the user's constraints and the assistant-side facts that may be needed for resolving follow-up references.

\begin{algorithm}[t]
\caption{Similarity-based window construction for multi-turn routing. The algorithm scans a dialogue from left to right, starts a new semantic window when adjacent user turns are weakly related, and otherwise extends the current window. For each current request, it outputs a context-enhanced prompt that contains only the active semantic window.}
\small
\label{alg:window-construction}
\KwIn{Dialogue $\mathcal{Z}_i=\{(u_{i,1},a_{i,1}),\ldots,(u_{i,m_i-1},a_{i,m_i-1}),u_{i,m_i}\}$; encoder $E(\cdot;w)$; threshold $\tau$}
\KwOut{Context-enhanced prompt $c_i$ and retained window $H_i$ for the current request $x_i=u_{i,m_i}$}

$H \gets \{u_{i,1}, a_{i,1}\}$\;
$e_{\mathrm{prev}} \gets E(u_{i,1};w)$\;
\For{$j=2$ \KwTo $m_i$}{
    $e_j \gets E(u_{i,j};w)$\;
    $\rho_j \gets \mathrm{sim}(e_j,e_{\mathrm{prev}})$\;
    \eIf{$\rho_j < \tau$}{
        $H \gets \{u_{i,j}\}$ \tcp*{topic shift; reset the semantic window}
    }{
        $H \gets H \cup \{u_{i,j}\}$ \tcp*{same local context; extend the window}
    }
    \If{$j < m_i$ \textnormal{and} $u_{i,j}\in H$}{
        $H \gets H \cup \{a_{i,j}\}$ \tcp*{retain assistant-side context inside the window}
    }
    $e_{\mathrm{prev}} \gets e_j$\;
}
$H_i \gets H$\;
$c_i \gets \mathrm{FormatPrompt}(H_i,x_i)$\;
\Return{$c_i,H_i$}\;
\end{algorithm}

\subsection{Routing and Training}
\label{sec:method:training}

For a query $x_i$, after being transformed into a combined query $c_i$ using the similarity-based window partitioning algorithm, \sys generates a selection probability distribution over $T$ candidate LLMs:
\begin{equation}
R(c_i; \theta)=
\mathrm{softmax}
\left(
\begin{bmatrix}
\mathrm{sim}(E(c_i; w), k_1)\\
\mathrm{sim}(E(c_i; w), k_2)\\
\vdots\\
\mathrm{sim}(E(c_i; w), k_T)
\end{bmatrix}
\right),
\label{eq:routing}
\end{equation}
where $\theta \equiv \{w, k_1, k_2, \ldots, k_T\}$ denotes the parameters of \sys, and $\mathrm{sim}(\cdot,\cdot)$ denotes cosine similarity.

A straightforward way to train the router is to align its output distribution with a score distribution:
\begin{equation}
\min_{\theta}
\sum_{(c_i,y_i)\in \mathcal{C}_{\text{train}}}
\mathrm{KL}
\Big(
R(c_i;\theta),\;
\mathrm{softmax}[s_i^{(1)},\ldots,s_i^{(T)}]
\Big),
\label{eq:kl_loss}
\end{equation}
where $\mathrm{KL}(\cdot,\cdot)$ denotes the Kullback--Leibler divergence~\cite{kullback1951information}.
This KL objective has recently been used in LLM routing~\cite{lu2023routing}.
However, we argue that it is not an ideal proxy for router training in our setting, because the goal of routing is to assign queries to top-performing LLMs rather than to fit the full score distribution, especially for bottom-performing models.
Following the contrastive learning spirit of RouterDC~\cite{chen2024RouterDC}, \sys adopts contrastive training objectives over the context-enhanced query.

\subsubsection{Sample-LLM Contrastive Loss}

For each combined query $c_i$, \sys constructs positive and negative LLM index sets $I_i^+$ and $I_i^-$.
The sample-LLM contrastive loss is defined as
\begin{equation}
L_{\text{sample-LLM}}(c_i,y_i;\theta)
=
-\log \frac{P_i^+}{P_i^+ + P_i^-},
\label{eq:sample_llm}
\end{equation}
where
\begin{equation}
P_i^+ = \sum_{t \in I_i^+} e^{\mathrm{sim}(E(c_i; w), k_t)},
\label{eq:p_pos}
\end{equation}
is the sum of the exponentiated similarities between the query embedding $E(c_i; w)$ and the top-$K_+$ model embeddings, and
\begin{equation}
P_i^- = \sum_{t \in I_i^-} e^{\mathrm{sim}(E(c_i; w), k_t)},
\label{eq:p_neg}
\end{equation}
is the corresponding sum over the bottom-$K_-$ model embeddings.
Here, $I_i^+$ and $I_i^-$ denote the index sets of the top-$K_+$ and bottom-$K_-$ LLMs, respectively.

\subsubsection{Sample-Sample Contrastive Loss}

In addition to contrasting candidate models for each query, \sys leverages unsupervised clustering to group semantically related queries and constructs a sample-sample contrastive loss:
This representation-learning design is related to scalable log-linear optimization and predictive structure learning across related tasks~\cite{andrew2007scalable,Ando2005}.
\begin{equation}
L_{\text{sample-sample}}(c_i;\theta)
=
-\log \frac{Q_i^+}{Q_i^+ + Q_i^-},
\label{eq:sample_sample}
\end{equation}
where
\begin{equation}
Q_i^+ = e^{\mathrm{sim}(E(c_i; w), E(c_i^+; w))}
\label{eq:q_pos}
\end{equation}
represents the similarity between the embedding of query $c_i$ and that of a randomly chosen in-group query $c_i^+$, and
\begin{equation}
Q_i^- = \sum_{c_i^- \in X_i^-} e^{\mathrm{sim}(E(c_i; w), E(c_i^-; w))}
\label{eq:q_neg}
\end{equation}
denotes the summed similarities between $c_i$ and a set of out-group queries $X_i^-$.

The overall training objective combines the above two contrastive losses:
\begin{equation}
\min_{\theta}
\sum
\left(
\alpha L_{\text{sample-LLM}}
+
\beta L_{\text{sample-sample}}
\right),
\label{eq:training_obj}
\end{equation}
where $\alpha$ and $\beta$ are hyperparameters that balance the two loss terms.

Algorithm~\ref{alg:training} summarizes the training and inference pipeline of \sys.
The algorithm uses Algorithm~\ref{alg:window-construction} as a preprocessing step to transform raw multi-turn dialogues into context-enhanced prompts.
Training then relies on model-level and sample-level contrastive supervision, while inference only requires a single router forward pass before calling the selected candidate LLM.

\begin{algorithm}[t]
\caption{Router training and inference for \sys. The procedure first converts raw dialogues into context-enhanced prompts using Algorithm~\ref{alg:window-construction}, then builds contrastive supervision from candidate-model scores, optimizes the router with sample-LLM and sample-sample losses, and finally selects one candidate LLM for each test prompt.}
\small
\label{alg:training}
\KwIn{Training dialogues $\mathcal{D}_{\text{train}}$, test dialogues $\mathcal{D}_{\text{test}}$, LLM pool $\mathcal{M}=\{M_t\}_{t=1}^{T}$, encoder $E(\cdot;w)$, model embeddings $\{k_t\}_{t=1}^{T}$, threshold $\tau$, hyperparameters $K_+$, $K_-$, $N$, $\lambda$, $b$, $\eta$}
\KwOut{Trained router $R(\cdot;\theta)$ and routed responses on the test set}

\textbf{Data preprocessing:}\\
\For{$\mathcal{Z}_i \in \mathcal{D}_{\text{train}}$}{
    $(c_i,H_i) \gets \mathrm{WindowConstruct}(\mathcal{Z}_i,E,\tau)$ using Algorithm~\ref{alg:window-construction}\;
    \For{$t=1$ \KwTo $T$}{
        Generate response $r_{i,t}\gets M_t(c_i)$\;
        Score $r_{i,t}$ with the judge model to obtain $w_{i,t}$\;
    }
}
$\mathcal{C}_{\text{train}} \gets \{(c_i,\{w_{i,t}\}_{t=1}^{T})\}$\;

\textbf{Training:}\\
Cluster queries $\{c_i\}_{i=1}^{n}$ into $N$ groups\\
\Repeat{converged}{
    Sample mini-batch $B$ from $\mathcal{C}_{\text{train}}$\\
    \For{$(c_i,y_i)\in B$}{
        Construct $I_i^+$ and $I_i^-$ using $w_{i,t}$\\
        Compute $L_{\text{sample-LLM}}(c_i,y_i;\theta)$\\
        Sample in-group query $c_i^+$ and out-group queries $\{c_i^-\}$\\
        Compute $L_{\text{sample-sample}}(c_i;\theta)$
    }
    $L(B;\theta)\gets \sum_{(c_i,y_i)\in B}\big[L_{\text{sample-LLM}}+\lambda L_{\text{sample-sample}}\big]$\\
    $\theta \gets \theta - \eta \nabla_{\theta} L(B;\theta)$
}

\textbf{Inference:}\\
\For{$\mathcal{Z}_i \in \mathcal{D}_{\text{test}}$}{
    $(c_i,H_i) \gets \mathrm{WindowConstruct}(\mathcal{Z}_i,E,\tau)$\;
    $\hat{t}_i \gets \arg\max_t R_t(c_i;\theta)$\;
    $\hat{y}_i \gets M_{\hat{t}_i}(c_i)$\;
}
\Return{$R(\cdot;\theta)$ and $\{\hat{y}_i\}$}\;
\end{algorithm}

\subsection{Decoupled Evaluation Metrics}
\label{sec:method:evaluation}

For each combined query $c_i$ and each candidate model $\mathcal{M}_t$ ($t=1,\ldots,T$), we obtain a \emph{true score} $0 \le w_{i,t} \le 1$ from judge models, which reflects the absolute response quality of $\mathcal{M}_t$ on $c_i$.

\paragraph{Evaluation Accuracy (absolute quality).}
Given the router's selection indicator $s_i^{(t)} \in \{0,1\}$, where $s_i^{(t)}=1$ if the router selects $\mathcal{M}_t$ for $c_i$ and $0$ otherwise, the score of the selected model is
\begin{equation}
s_i = \sum_{t=1}^{T} w_{i,t}\cdot s_i^{(t)}.
\label{eq:si}
\end{equation}
We define \textit{evaluation accuracy} as the average true score achieved by the routing system:
\begin{equation}
\overline{si}
=
\frac{1}{n}\sum_{i=1}^{n} s_i,
\label{eq:acc_eval}
\end{equation}
which reflects the absolute usefulness of the routed responses under the constructed prompts.

\paragraph{Construction Accuracy (prompt quality).}
To characterize the overall quality of the constructed prompt across all candidate models, we define the average true score
\begin{equation}
\bar{w}
=
\frac{1}{nT}\sum_{i=1}^{n}\sum_{t=1}^{T} w_{i,t},
\label{eq:mean_w}
\end{equation}
which reflects the average capability of the candidate model pool under a given prompt construction. A higher $\bar{w}$ indicates that the constructed prompt enables better responses across all candidate models.

\paragraph{Router Performance (relative improvement).}
Based on $\overline{si}$ and $\bar{w}$, we define router performance as
\begin{equation}
P_{\text{router}}
=
\frac{\overline{si}}{\bar{w}}.
\label{eq:p_router}
\end{equation}
Here, $\bar{w}$ reflects the average capability of the candidate model pool under a given prompt construction, while $P_{\text{router}}$ measures how much the router improves over this average by selecting stronger models. A value greater than 1 indicates that the router outperforms random model selection.

\begin{figure}[t]
    \centering
    \includegraphics[width=0.98\linewidth]{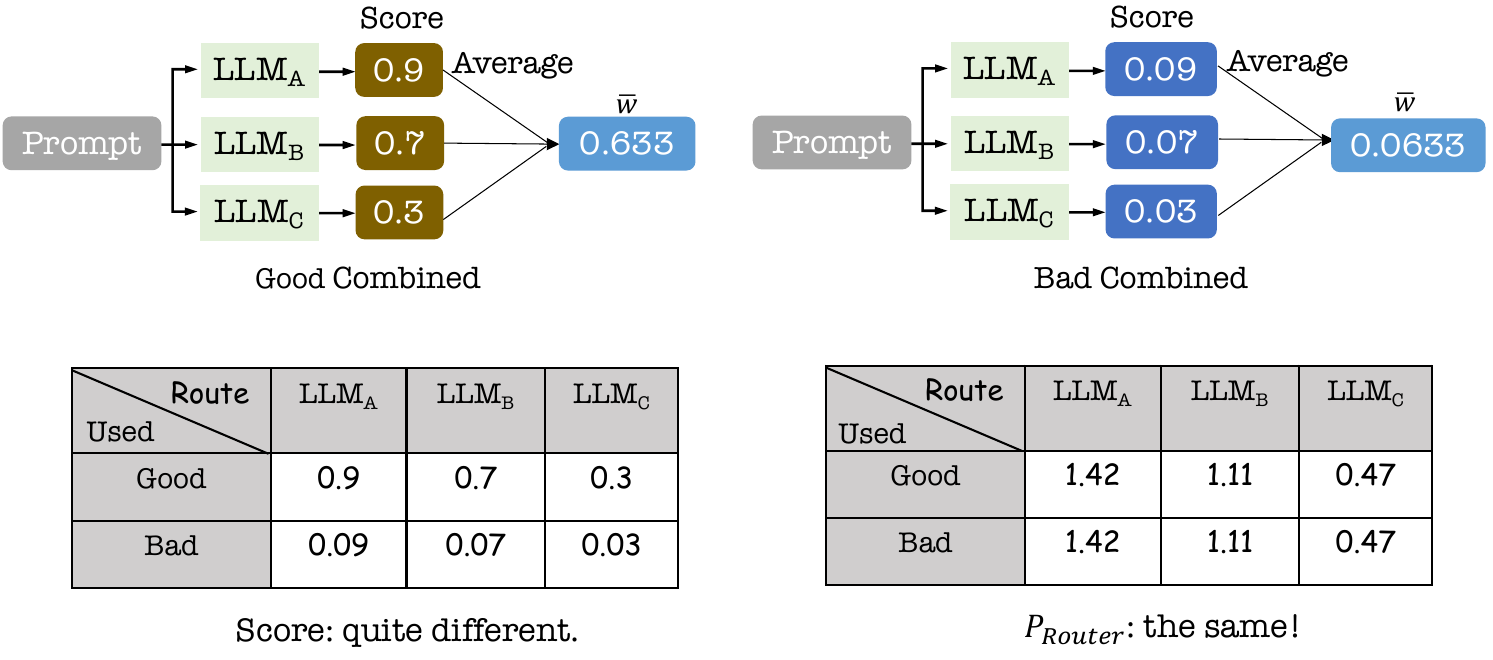}
    \caption{
    Illustration of the three complementary metrics used for decoupled evaluation.
    For each constructed prompt, all candidate LLMs are first evaluated with absolute true scores.
    Construction accuracy $\bar{w}$ averages these scores across the whole candidate pool and therefore measures whether the prompt construction stage preserves useful context for models in general.
    Evaluation accuracy $\overline{si}$ then records the true score of the single model selected by the router, representing the end-to-end quality observed by users.
    Router performance $P_{\text{router}}=\overline{si}/\bar{w}$ normalizes the selected-model score by the candidate average, showing whether the router improves over random model selection under the same constructed prompts.
    }
    \Description{Diagram comparing construction accuracy, evaluation accuracy, and router performance for multi-turn routing.}
    \label{fig:three_metrics}
\end{figure}

\paragraph{Discussion: three complementary metrics.}
For evaluation in multi-turn dialogue, we adopt three complementary metrics to characterize the system: \textit{evaluation accuracy} ($\overline{si}$), \textit{construction accuracy} ($\bar{w}$), and \textit{router performance} ($P_{\text{router}}$).
Figure~\ref{fig:three_metrics} illustrates why these metrics are complementary.
Under good prompt construction, the candidate models obtain scores of $\{0.9, 0.7, 0.3\}$, yielding an average construction accuracy of $\bar{w}=0.633$.
If the router selects the best model, then $\overline{si}=0.9$ and $P_{\text{router}}=0.9/0.633=1.42$, indicating both strong construction quality and effective routing.
In contrast, under poor prompt construction, information loss uniformly degrades the candidate scores to $\{0.09, 0.07, 0.03\}$, reducing construction accuracy to $\bar{w}=0.0633$.
Even if the router still selects the best model, the final evaluation accuracy drops to $\overline{si}=0.09$, while the router performance remains unchanged: $P_{\text{router}}=0.09/0.0633=1.42$.
This example shows that high router performance does not necessarily imply high end-to-end response quality.
Instead, $\overline{si}$ reflects the final usefulness of the routed response, $\bar{w}$ measures the intrinsic quality of prompt construction, and $P_{\text{router}}$ quantifies the router's ability to select better-than-average candidates.
Together, these metrics enable a decoupled analysis of construction quality and routing effectiveness.

\section{Implementation}
\label{sec:implementation}

We implement \sys following the standard router-training and evaluation pipeline for multi-turn LLM routing. Our implementation consists of candidate model construction, multi-turn dataset preprocessing, similarity-window segmentation, response scoring, and router optimization.

\textbf{Candidate LLMs.}
We evaluate \sys on seven open-source LLMs from HuggingFace: Mistral-7B~\cite{jiang2023mistral}, MetaMath-Mistral-7B~\cite{yu2023metamath}, zephyr-7b-beta~\cite{tunstall2023zephyr}, Chinese-Mistral-7B~\cite{Chinese-Mistral}, dolphin-2.6-mistral-7b~\cite{dolphin-2.6-mistral-7b}, Llama-3-8B~\cite{dubey2024llama}, and dolphin-2.9-llama3-8b~\cite{dolphin-2.9-llama3-8b}. The first five models are Mistral-based, while the last two are Llama-3-based. These models cover general-purpose, instruction-tuned, domain-tuned, and language-adapted variants, enabling us to evaluate whether \sys can select among LLMs with heterogeneous capabilities.
The Llama-family candidates build on the broader open-foundation-model line represented by LLaMA and LLaMA 2~\cite{touvron2023llama,touvron2023llama2}, and zephyr-7b-beta follows preference-optimization techniques such as direct preference optimization~\cite{rafailov2024direct}.

\textbf{Datasets.}
We conduct experiments on two multi-turn dialogue datasets, MTBench~\cite{bai2024mt} and ShareGPT~\cite{sharegpt_raw}. MTBench contains multi-turn conversations across diverse task categories, while ShareGPT consists of real user-AI interactions. We preprocess ShareGPT by removing noisy samples and filtering out non-multi-turn dialogues. For both datasets, we randomly split the data into 70\% for training and 30\% for testing. All training samples are combined as $\mathcal{C}_{\text{train}}$ for router training.

\textbf{Baselines.}
We compare \sys with three groups of baselines. First, \textbf{Single Model} directly uses one candidate LLM to answer each query without routing. This group includes all seven candidate LLMs listed above. Second, \textbf{Conv-ID Context} preserves the original dialogue segmentation according to conversation IDs and performs routing with the full original multi-turn context. This baseline serves as an ID-based static context reference. Third, \textbf{ZOOTER}~\cite{lu2023routing} is used as a strong single-turn router backbone. To evaluate the generality of the proposed similarity-window mechanism, we combine ZOOTER with the same similarity-window segmentation and compare it with \sys.

\textbf{Evaluation protocol.}
We use the Language Model Evaluation Harness~\cite{eval-harness} for model evaluation, following common benchmark practice from MMLU-style multitask evaluation~\cite{mmlu}. For open-ended generation, we generate $M=10$ candidate responses using stochastic beam search, where beam expansion is sampled with temperature 0.2. The generated responses are then scored by judge models, and the router is evaluated using testing weighted accuracy.

\textbf{Similarity-window configurations.}
We use mDeBERTaV3-base~\cite{he2021deberta} as the encoder $E(x;w)$, which contains approximately 86M backbone parameters. This choice follows the use of efficient neural encoders for structured text analysis~\cite{rasooli-tetrault-2015}. The similarity threshold is set to $\tau=0.91$ for constructing similarity windows. The LLM embedding dimension is 768, and the number of clusters is set to $N=5$.

\textbf{Router training configurations.}
The router is trained for 1,000 steps using AdamW~\cite{loshchilov2018decoupled}. The learning rate is $5 \times 10^{-5}$, the weight decay is 0.01, and the batch size is 16. All experiments are conducted on NVIDIA A100 80GB GPUs.

\section{Evaluation}
\label{sec:evaluation}

We evaluate \sys to answer the following questions:
\begin{enumerate}[left=0.5em]
    \item Can \sys improve routing performance on multi-turn dialogue tasks?
    \item How well does \sys generalize to out-of-distribution (OOD) scenarios?
    \item How much does each component contribute to the effectiveness of \sys?
    \item How sensitive is \sys to key hyperparameters?
    \item What is the computational overhead introduced by \sys?
\end{enumerate}

\subsection{Evaluation Setup}

\subsubsection{Prompts Used for Evaluation}

To evaluate the quality of routed responses in multi-turn dialogue, we employ a judge model with the following evaluation prompt:
The prompt is written as a structured evaluation rubric following standard academic reporting conventions~\cite{APA:83}.

\begin{quote}
\textit{You are given a multi-turn conversation, including the previous dialogue history, the current user request, and an assistant response generated by a candidate LLM. Please evaluate the quality of the assistant response with a single integer score from 1 to 10. Judge the response semantically rather than by exact wording: paraphrases, different ordering, or different surface expressions should be treated as acceptable when they preserve the same meaning and satisfy the user's request. Do not reward verbosity by itself; longer answers should receive higher scores only when the additional content improves correctness, usefulness, or coverage.}

\textit{Use the following criteria when assigning the score. \textbf{Intent satisfaction} evaluates whether the response identifies the user's actual goal in the current turn and provides a useful answer or action for that goal. \textbf{Dialogue context usage} evaluates whether the response correctly incorporates relevant prior turns, carries over established constraints, resolves references such as ``it'' or ``that'' appropriately, and avoids being distracted by obsolete or unrelated context. \textbf{Instruction following} evaluates whether the response obeys explicit requirements about format, language, style, scope, tools, refusal behavior, or output length. \textbf{Factuality and reasoning} evaluates whether the response is factually correct, logically consistent, and supported by the information available in the conversation or by reliable reasoning. \textbf{Completeness} evaluates whether all important parts of the request are addressed, including edge cases, caveats, or requested explanations. \textbf{Clarity and usefulness} evaluates whether the response is easy to understand, well organized, appropriately concise, and practically helpful to the user.}

\textit{Apply the following scoring guideline. A score of \textbf{10} should be reserved for a response that fully satisfies the user's intent, uses all necessary dialogue context correctly, follows every instruction, is factually sound, complete, and clear. A score of \textbf{9} indicates an excellent response with only negligible wording or presentation issues. Scores of \textbf{7--8} indicate a mostly correct and useful response that satisfies the main request but has minor omissions, mild ambiguity, limited explanation, or only partial use of relevant context. Scores of \textbf{5--6} indicate a partially useful response that addresses the general topic but misses important constraints, overlooks meaningful dialogue history, contains noticeable but non-fatal factual or reasoning errors, or leaves substantial parts of the task incomplete. Scores of \textbf{3--4} indicate a weak response that is only loosely related to the request, substantially incomplete, poorly grounded in the dialogue, or affected by major factual, reasoning, or instruction-following errors. Scores of \textbf{1--2} indicate a response that is irrelevant, contradicts the conversation, refuses without justification, is unsafe when safety is required, fails to answer the user, or is dominated by hallucinated content.}

\textit{When several issues are present, choose the score that best reflects the overall usefulness of the response to the user. Penalize strongly for errors that would cause the user to take a wrong action, violate an explicit constraint, or misunderstand the task outcome. A response with a major factual error should not receive a score above 6, even if it is fluent. A response that ignores the current user request or relies on the wrong dialogue context should not receive a score above 4. A response that is correct but unnecessarily verbose, poorly organized, or missing minor details may still receive a high score if the user's intent is satisfied.}

\textit{Return only the final score.}
\end{quote}

The judge model returns an integer rating from 1 to 10, which we normalize to the range $[0, 1]$ as the true score $w_{i,t}$ for evaluation metrics computation.

\subsection{Overall Performance}

\begin{figure}[t]
    \centering
    \includegraphics[width=0.96\linewidth]{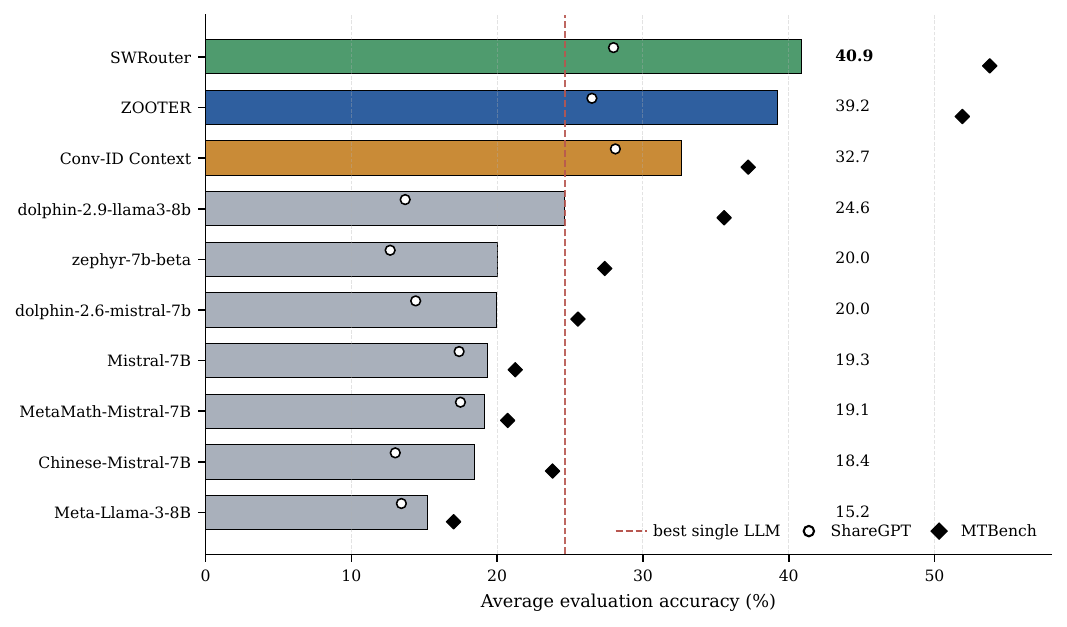}
    \caption{
    Overall \textbf{evaluation accuracy} (\%) on ShareGPT and MTBench.
    Horizontal bars rank all candidate single LLMs and routing methods by average accuracy, while markers show the corresponding ShareGPT and MTBench scores.
    The dashed line marks the strongest single LLM, making the routing gain of \sys visually explicit.}
    \Description{Ranked horizontal bar chart showing average evaluation accuracy for all single LLMs and routing methods, with markers for ShareGPT and MTBench scores.}
    \label{fig:main_results}
\end{figure}

As shown in Figure~\ref{fig:main_results}, \sys achieves the best overall performance across the two multi-turn dialogue datasets. Compared with the strongest single LLM, dolphin-2.9-llama3-8b, \sys improves the average \textbf{evaluation accuracy} from 24.63\% to 40.89\%, corresponding to an absolute gain of 16.26\%. The improvement is consistent across both datasets, with gains of 14.29\% on ShareGPT and 18.22\% on MTBench.

Compared with Conv-ID Context, which preserves the original dialogue segmentation by conversation IDs, \sys further improves the average \textbf{evaluation accuracy} by 8.22\%. This result suggests that exact ID-based partitioning is not necessarily the optimal context construction strategy for multi-turn routing. Instead, dynamically constructing semantic windows provides more useful routing inputs.

Moreover, ZOOTER equipped with the similarity-window mechanism achieves 39.21\% average \textbf{evaluation accuracy}, substantially outperforming Conv-ID Context. This indicates that the proposed similarity-window design is general and can improve different routing backbones by constructing better multi-turn contexts.

\subsection{OOD Generalization}

\begin{figure}[t]
    \centering
    \includegraphics[width=0.9\linewidth]{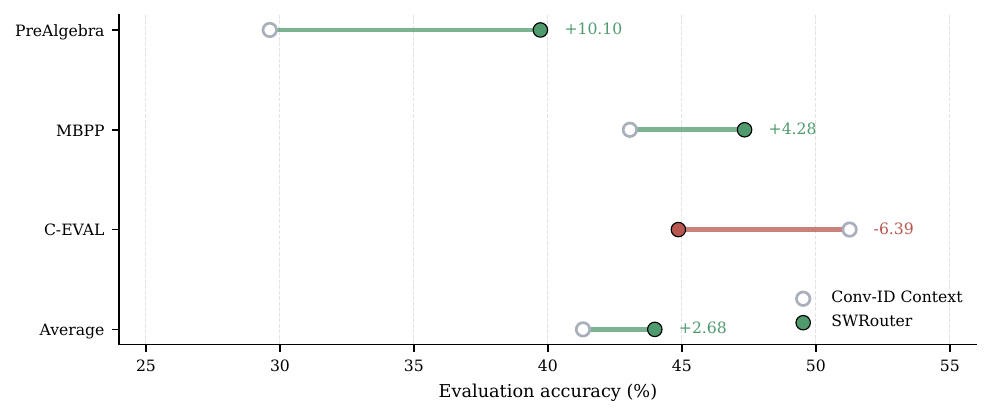}
    \caption{
    \textbf{Evaluation accuracy} (\%) of \sys and Conv-ID Context on out-of-distribution datasets.
    PreAlgebra, MBPP, and C-EVAL cover mathematical reasoning, code generation, and Chinese knowledge-intensive evaluation, respectively.
    The connected markers compare \sys with Conv-ID Context on each task, and the right-side annotations report the absolute gain or loss.
    This visualization emphasizes that \sys improves the average OOD result despite not winning on every individual task.}
    \Description{Dumbbell chart comparing SWRouter and Conv-ID Context on PreAlgebra, MBPP, C-EVAL, and their average, with gain annotations.}
    \label{fig:ood_results}
\end{figure}

We further evaluate the generalization ability of \sys under out-of-distribution scenarios, including \textit{PreAlgebra}, \textit{MBPP}, and \textit{C-EVAL}. These datasets cover mathematical reasoning, code generation, and Chinese knowledge-intensive evaluation, respectively, and are related to established mathematical and Chinese multitask evaluation settings~\cite{gsm8k,cmmlu}.

As shown in Figure~\ref{fig:ood_results}, \sys achieves the best average OOD performance, reaching 43.99\% and outperforming Conv-ID Context by 2.68\%. In particular, \sys improves Conv-ID Context by 10.10\% on PreAlgebra and 4.28\% on MBPP. Although Conv-ID Context performs better on C-EVAL, \sys exhibits stronger average robustness across heterogeneous OOD tasks.

This result indicates that \sys does not simply fit ID-based routing labels. Instead, the learned selection strategy captures transferable routing patterns that generalize beyond the original dialogue distribution.

\subsection{Performance Breakdown}
\label{sec:eval:breakdown}

\begin{figure}[t]
    \centering
    \includegraphics[width=0.9\linewidth]{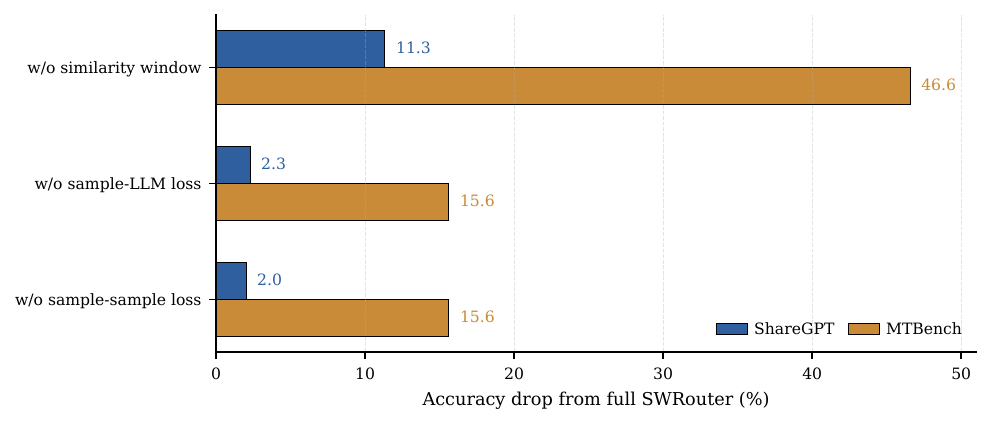}
    \caption{
    Ablation study results measured by \textbf{evaluation accuracy} (\%) of \sys.
    Each bar reports the accuracy drop relative to the full \sys model after removing one component.
    The full model obtains 27.98\% on ShareGPT and 53.79\% on MTBench.
    This drop-oriented view highlights that similarity-window construction is the dominant contributor, especially on MTBench, while both contrastive losses also provide consistent gains.}
    \Description{Bar chart showing accuracy drops from full SWRouter after removing similarity-window construction, sample-LLM loss, or sample-sample loss.}
    \label{fig:ablation_accuracy}
\end{figure}

We conduct ablation studies to quantify the contribution of each component in \sys. As shown in Figure~\ref{fig:ablation_accuracy}, removing any component consistently degrades performance, showing that all components are necessary for effective routing.

\textbf{Similarity window.}
Removing the similarity window causes the largest performance drop. \textbf{Evaluation Accuracy} decreases by 11.33\% on ShareGPT and 46.57\% on MTBench. This confirms that adaptive context construction is the most important component of \sys.

\textbf{Sample-LLM loss.}
Removing $L_{\text{sample-LLM}}$ reduces performance to 25.70\% on ShareGPT and 38.17\% on MTBench. This indicates that modeling the relationship between dialogue samples and candidate LLMs is critical for accurate routing.

\textbf{Sample-sample loss.}
Removing $L_{\text{sample-sample}}$ also significantly hurts performance, especially on MTBench. This shows that sample-level representation alignment helps construct a more structured routing space.

\textbf{Actual score.}
We further observe that proxy routing scores can substantially overestimate the final end-to-end performance if prompt construction is ignored. This highlights the necessity of evaluating actual generated responses rather than relying only on intermediate routing objectives.

\subsection{Sensitivity Analysis}
\label{sec:sensitivity}

\begin{figure}[t]
    \centering
    \begin{subfigure}[t]{0.48\linewidth}
        \centering
        \includegraphics[width=\linewidth]{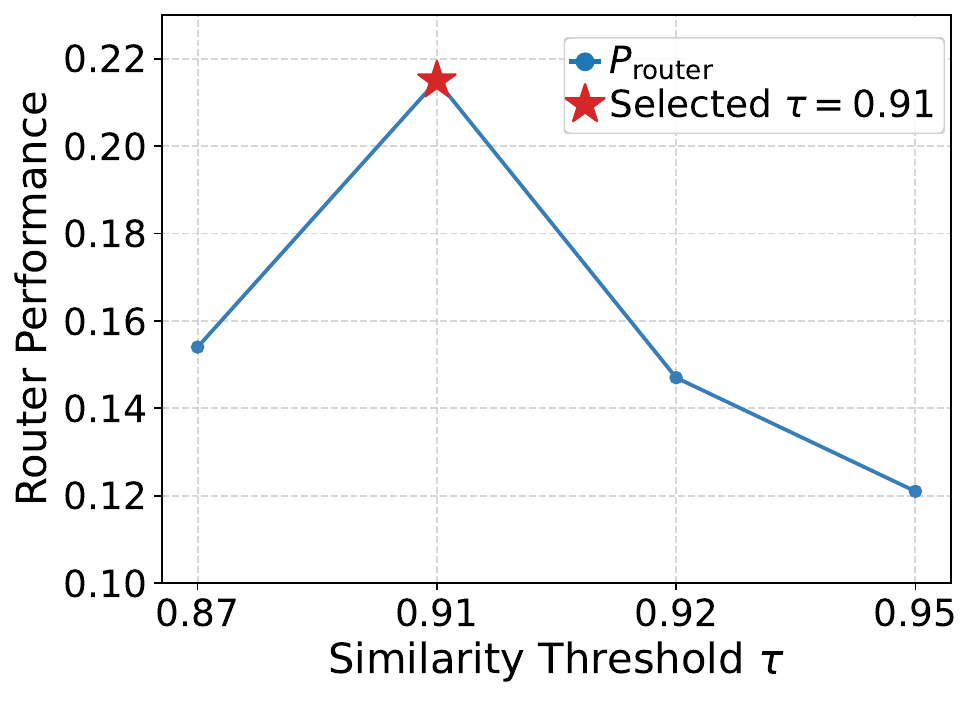}
        \caption{Router performance $P_{\text{router}}$.}
        \label{fig:router_performance}
    \end{subfigure}%
    \hfill
    \begin{subfigure}[t]{0.48\linewidth}
        \centering
        \includegraphics[width=\linewidth]{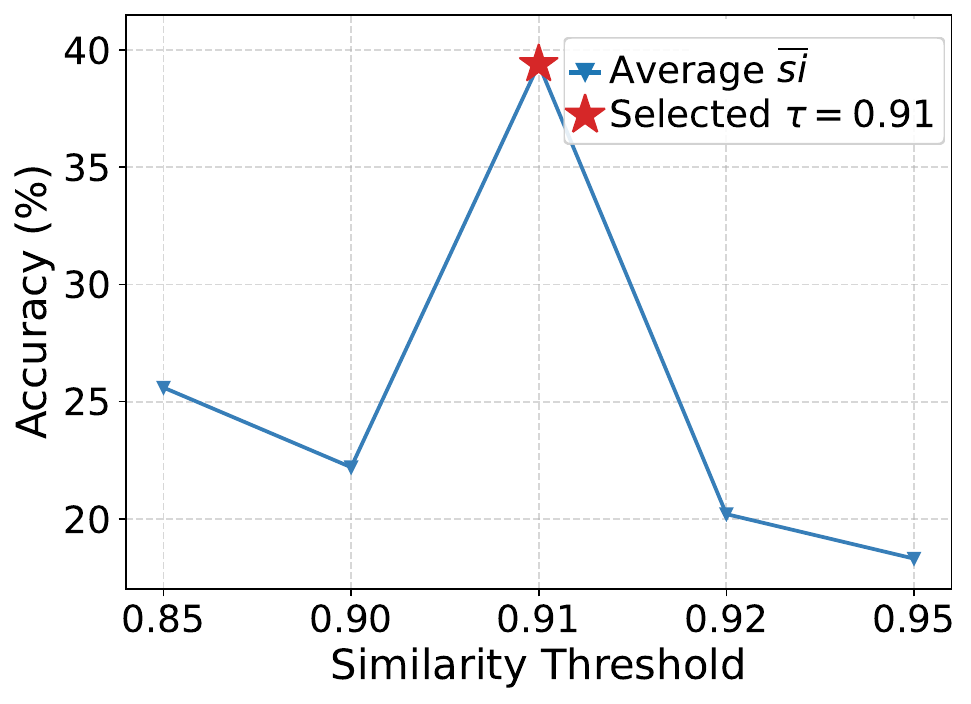}
        \caption{Average value of $\overline{si}$.}
        \label{fig:average_si}
    \end{subfigure}
    \caption{Effect of the similarity threshold $\tau$ on router performance $P_{\text{router}}$ and the average value of $\overline{si}$.
    The left subfigure shows how strongly the router improves over the average candidate-model score under each threshold, while the right subfigure reports the resulting end-to-end routed response quality.
    Together, the curves show that threshold selection affects both model-selection effectiveness and absolute response quality, with the best operating region concentrated around $\tau=0.91$.}
    \Description{Two line charts showing how the similarity threshold affects router performance and average evaluation accuracy.}
    \label{fig:router_and_average_si}
\end{figure}

\begin{figure}[t]
    \centering
    \begin{subfigure}[t]{0.48\linewidth}
        \centering
        \includegraphics[width=\linewidth]{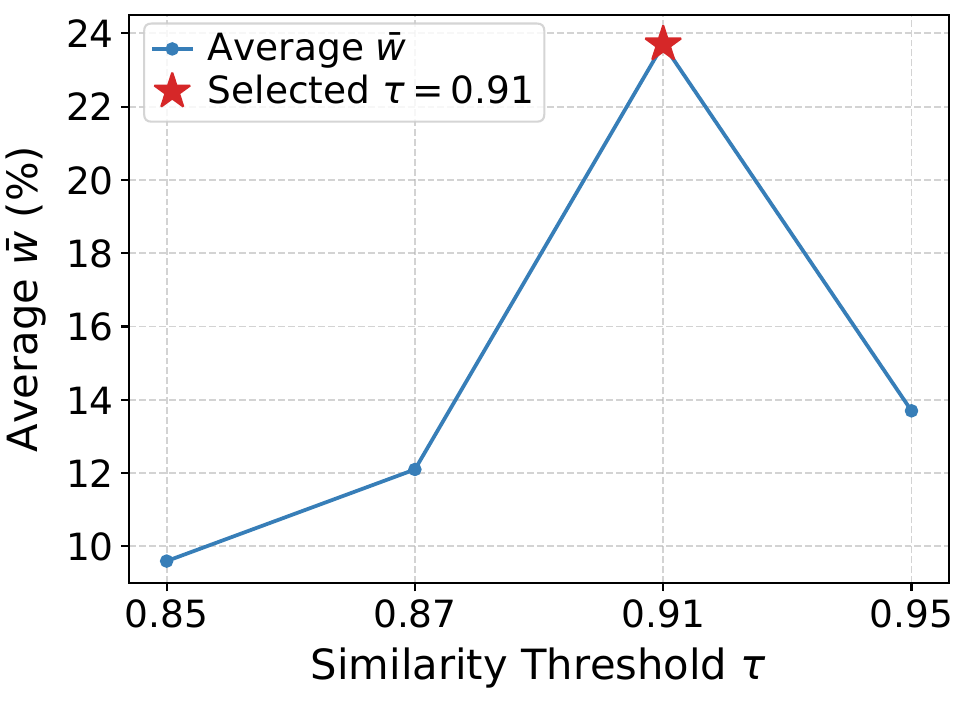}
        \caption{MT-Bench.}
        \label{fig:mtbench_wbar}
    \end{subfigure}%
    \hfill
    \begin{subfigure}[t]{0.48\linewidth}
        \centering
        \includegraphics[width=\linewidth]{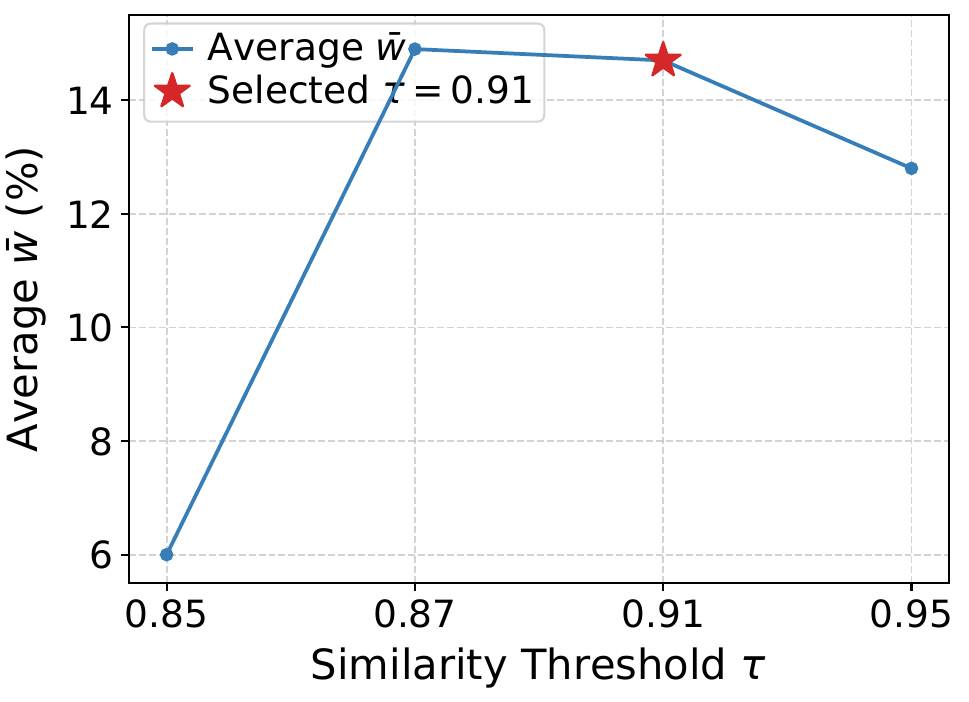}
        \caption{ShareGPT.}
        \label{fig:sharegpt_wbar}
    \end{subfigure}
    \caption{Effect of the similarity threshold $\tau$ on the average value of $\bar{w}$ for MT-Bench and ShareGPT.
    Since $\bar{w}$ averages the true scores of all candidate LLMs under the constructed prompts, these curves isolate the prompt-construction quality from the router's model-selection behavior.
    The comparison shows how stricter or looser window partitioning changes the amount of useful context retained for the candidate model pool.}
    \Description{Two line charts showing how the similarity threshold affects construction accuracy on MT-Bench and ShareGPT.}
    \label{fig:mtbench_sharegpt_wbar}
\end{figure}

We analyze the sensitivity of \sys to the similarity threshold $\tau$.

\textbf{Effect of Similarity Threshold $\tau$.}We conduct a sensitivity study on the similarity threshold $\tau$ with respect to three key performance metrics: the average value of $\overline{si}$, the average value of $\bar{w}$, and router performance $P_{\text{router}}$.

\textbf{Average Value of $\overline{si}$}
($\overline{si}=\frac{1}{n} \sum_{i=1}^n s_i$).
Figure~\ref{fig:router_and_average_si}(b) shows the effect of $\tau$ on $\overline{si}$. The similarity-window construction achieves its best performance at $\tau=0.91$, reaching 40.89\% accuracy and outperforming the Conv-ID Context baseline of 32.67\%. When $\tau$ moves away from this value, the performance decreases noticeably. For example, the accuracy drops to 25.6\% at $\tau=0.85$, 22.2\% at $\tau=0.90$, 20.2\% at $\tau=0.92$, and 18.3\% at $\tau=0.95$. These results indicate that $\overline{si}$ is highly sensitive to the similarity threshold, and that $\tau=0.91$ provides the most effective similarity window.

\textbf{Average Value of $\bar{w}$.}
Figure~\ref{fig:mtbench_sharegpt_wbar} presents the average value of $\bar{w}$ under different similarity thresholds $\tau$ on MT-Bench and ShareGPT. The results show that $\bar{w}$ is strongly affected by the similarity threshold. On MT-Bench, $\bar{w}$ increases from 9.60\% at $\tau=0.85$ to 12.10\% at $\tau=0.87$, and further reaches the maximum value of 23.70\% at $\tau=0.91$. However, when the threshold is increased to $\tau=0.95$, $\bar{w}$ drops sharply to 13.70\%, suggesting that an overly strict similarity constraint can substantially reduce the retained high-quality samples.

A similar trend can also be observed on ShareGPT. The average value of $\bar{w}$ is 6.00\% at $\tau=0.85$, increases significantly to 14.90\% at $\tau=0.87$, and remains at a comparable level of 14.70\% when $\tau=0.91$. It then decreases to 12.80\% at $\tau=0.95$. These results suggest that ShareGPT is slightly more stable than MT-Bench around the optimal region, but its performance is still affected by threshold selection.

Overall, the high-quality region is mainly concentrated around $\tau=0.91$. MT-Bench achieves its best result at $\tau=0.91$, while ShareGPT maintains competitive performance in the range of $\tau \in [0.87, 0.91]$. Considering both datasets, $\tau=0.91$ provides a stable trade-off between sample quality and coverage, and we therefore adopt it as the default threshold.

\textbf{Router Performance} ($P_{\text{router}}$).
Figure~\ref{fig:router_and_average_si}(a) illustrates the effect of the similarity threshold $\tau$ on router performance $P_{\text{router}}$. Overall, $P_{\text{router}}$ is influenced by the choice of $\tau$, although its variation is less pronounced than that of $\bar{w}$. Specifically, $P_{\text{router}}$ is 15.4\% at $\tau=0.87$, increases to its peak of 21.5\% at $\tau=0.91$, and then decreases to 14.7\% at $\tau=0.92$ and 12.1\% at $\tau=0.95$.

These results indicate that router performance still depends on the similarity threshold, but the overall fluctuation is relatively moderate. Compared with $\bar{w}$, $P_{\text{router}}$ is less sensitive to changes in $\tau$. Nevertheless, the best router performance is still achieved at $\tau=0.91$, further supporting our choice of $\tau=0.91$ as the default threshold.

\subsection{Robustness Analysis}

\subsubsection{Extension: Top-K Retrieval for Long-Range Dependency Recovery}

As an extension to the core similarity-based windowing approach, we investigate whether augmenting it with a lightweight Top-K retrieval mechanism can further recover semantically related queries that are separated by unrelated turns.

We simulate a scenario where a user issues a query \textit{a}, then several unrelated queries, and finally a related query \textit{b}. For each \textit{b}, similarity-based Top-K retrieval is performed over the dialogue history to test whether \textit{a} can be recalled.

\begin{table}[!htbp]
\small
\centering
\caption{Recall performance of similarity-based retrieval under different levels of semantic interference.
Inserted queries denote unrelated turns placed between two semantically related user requests; Top-$K$ recall measures whether the retrieval module can recover the earlier related turn from the dialogue history.}
\label{tab:retrieval_regular}
\begin{tabular}{lccc}
\toprule
\textbf{Inserted Queries} & \textbf{Top-1 Recall} & \textbf{Top-3 Recall} & \textbf{Top-5 Recall} \\
\midrule
1 & 79.43 & 100.00 & 100.00 \\
3 & 65.16 & 91.31 & 100.00 \\
5 & 59.84 & 77.84 & 94.15 \\
\bottomrule
\end{tabular}
\end{table}

We also test a more challenging case where 20 unrelated queries are inserted between \textit{a} and \textit{b}. The results are shown below.

\begin{table}[!htbp]
\small
\centering
\caption{Retrieval robustness under the high-difficulty setting with 20 distractor turns.
This setting stresses long-range dependency recovery by separating two related requests with many unrelated turns, and the reported recall indicates whether Top-$K$ retrieval can still recover the relevant earlier query.}
\label{tab:retrieval_hard}
\begin{tabular}{lccc}
\toprule
\textbf{Inserted Queries} & \textbf{Top-K} & \textbf{Recall (\%)} & \textbf{Difficulty} \\
\midrule
20 & 5 & 66.13 & Hard \\
\bottomrule
\end{tabular}
\end{table}

As shown in Tables~\ref{tab:retrieval_regular} and~\ref{tab:retrieval_hard}, similarity-based retrieval substantially improves robustness under semantic drift. Top-3 retrieval provides an effective trade-off between recall and efficiency. Even under high-difficulty conditions with 20 distractors, Top-5 recall remains above 66\%, indicating strong resilience. These results demonstrate that augmenting local similarity-based segmentation with retrieval mechanisms can further enhance context reconstruction in realistic multi-turn dialogue scenarios.

\subsubsection{Semantic Coherence in Long Dialogues}
\label{sec:semantic_coherence}

To further analyze the potential semantic drift as the dialogue length increases, we conducted a similarity analysis over multi-turn dialogues on the \textit{ShareGPT} and \textit{MT-Bench} datasets. Specifically, we computed the cosine similarity between the first and last user turns within each dialogue window under different minimum round sizes (20, 25, and 30).

\begin{table*}[t]
\small
\centering
\caption{Cosine similarity between the first and last user turns across dialogues of varying lengths.
For each dataset, we group dialogues by minimum round size and report the proportion of long-range turn pairs above or below the default similarity threshold $\tau=0.91$, together with the maximum, minimum, and average similarity.
The table characterizes how quickly semantic drift appears as dialogue length increases.}
\label{tab:semantic_drift}
\begin{tabular}{lccccccc}
\toprule
\textbf{Dataset} & \textbf{Min Group Size} & \textbf{Total Pairs} & \textbf{$\ge$0.91 (\%)} & \textbf{$<$0.91 (\%)} & \textbf{Max} & \textbf{Min} & \textbf{Avg} \\
\midrule
\textbf{ShareGPT} & 20 & 19 & 73.68 & 26.32 & 0.9922 & 0.6774 & 0.9283 \\
 & 25 & 13 & 76.92 & 23.08 & 0.9922 & 0.6774 & 0.9211 \\
 & 30 & 7 & 71.43 & 28.57 & 0.9874 & 0.8035 & 0.9247 \\
 \midrule
\textbf{MT-Bench} & 20 & 7 & 57.14 & 42.86 & 0.9620 & 0.7961 & 0.8989 \\
 & 25 & 6 & 50.00 & 50.00 & 0.9620 & 0.7961 & 0.8928 \\
 & 30 & 4 & 25.00 & 75.00 & 0.9426 & 0.7961 & 0.8603 \\
\bottomrule
\end{tabular}
\end{table*}

As shown in Table~\ref{tab:semantic_drift}, in \textit{ShareGPT}, over 70\% of long-range pairs maintain a cosine similarity above 0.91, and the average similarity remains above 0.92, suggesting strong semantic consistency even across extended dialogues. In \textit{MT-Bench}, the average similarity is slightly lower but still indicates limited semantic drift. Extremely low similarity cases ($<$0.80) are rare, implying that our similarity-based segmentation remains semantically coherent over long dialogue windows.

\subsection{Detailed Overhead Analysis}
\label{sec:overhead}

We conduct a comprehensive analysis of the computational overhead introduced by \sys from three aspects: inference latency, training cost, and token generation cost.

\subsubsection{Latency Analysis}
\label{sec:overhead:latency}

\begin{table}[!htbp]
\small
\centering
\caption{Time consumption of \sys during inference.
The table decomposes latency into similarity calculation for window construction, router selection, and downstream LLM inference, showing that the additional cost introduced by similarity-window routing is small relative to candidate-model generation.}
\label{tab:overhead_latency}
\resizebox{0.5\textwidth}{!}{
\begin{tabular}{lccc}
\toprule
\textbf{Time} & \textbf{Similarity Calculation} & \textbf{Router Selection} & \textbf{LLM Inference} \\
\midrule
\sys & 50.6s & 16min32s & $>$6h \\
\bottomrule
\end{tabular}
}
\end{table}

As shown in Table~\ref{tab:overhead_latency}, the similarity calculation in \sys accounts for less than 1\% (approximately 2.2‰) of the total inference time, indicating high computational efficiency. The dominant cost comes from LLM inference, which is consistent with all routing-based systems.

\subsubsection{Training Cost Analysis}
\label{sec:overhead:training}

The training process of \sys consists of four steps:

\begin{enumerate}
    \item \textbf{Dataset window partitioning}: We use a lightweight encoder (microsoft/mdeberta-v3-base) with approximately 86M backbone parameters, which is much smaller than the 7B/8B candidate LLMs, making this step's cost negligible.

    \item \textbf{Candidate response generation}: We use the seven small models mentioned in the paper (Mistral-7B, MetaMath-Mistral-7B, zephyr-7b-beta, Chinese-Mistral-7B, dolphin-2.6-mistral-7b, Meta-Llama-3-8B, dolphin-2.9-llama3-8b) to generate candidate answers for each query.

    \item \textbf{Judge-model scoring}: We use GPT for scoring. Since the output is just a single numerical score, this step does not significantly increase memory usage.

    \item \textbf{Router training}: The similarity-window encoder introduces limited overhead: microsoft/mdeberta-v3-base has approximately 86M backbone parameters, which is much smaller than the 7B/8B candidate LLMs. Moreover, the encoder is used only for lightweight similarity computation, while the dominant cost remains candidate LLM inference. Router training is performed only once before deployment, making it a one-time overhead.
\end{enumerate}

\subsubsection{Token Generation Cost}
\label{sec:overhead:token}

To evaluate the efficiency of model selection, we compare the token-level costs between \sys and the Conv-ID Context method. Due to the unavailability of precise pricing information for small-sized models such as 7B and 8B, we adopt token count as a unified proxy for cost estimation.

\begin{table}[!htbp]
\small
\centering
\caption{Overall evaluation accuracy and token cost comparison.
Evaluation accuracy measures response quality after routing, while token cost counts the generated tokens used by each routing strategy as a model-agnostic proxy for serving cost.
The comparison highlights the effectiveness--cost tradeoff between Conv-ID Context, SimWindow + ZOOTER, and \sys.}
\label{tab:router_avg_cost}
\begin{tabular}{lcc}
\toprule
\textbf{Router} & \textbf{Evaluation Acc.} $\uparrow$ & \textbf{Token Cost} $\downarrow$ \\
\midrule
Conv-ID Context & 32.67 & 2834 \\
SimWindow + ZOOTER & 39.20 & 3405 \\
\sys & \textbf{40.89} & 3477 \\
\bottomrule
\end{tabular}
\end{table}

As shown in Table~\ref{tab:router_avg_cost}, \sys incurs approximately $1.23\times$ the token cost of Conv-ID Context while achieving an $8.22\%$ absolute improvement in evaluation accuracy. Compared to ZOOTER, \sys requires only $1.02\times$ the token cost, yet delivers a further $1.69\%$ accuracy gain. This result suggests that \sys achieves a reasonable trade-off between performance and efficiency, despite lacking access to ground-truth optimal selections.

\subsubsection{Additional Analysis}

We further analyze the experimental results from the following four aspects:

\textbf{(1) Prompt Construction Impact.}
The similar performance patterns across models at different similarity thresholds suggest that prompt construction, rather than model architecture, is the primary performance driver. $\tau=0.91$ performs best among tested thresholds; performance is relatively strong in the neighborhood around 0.91.

\textbf{(2) Router-Score Correlation.}
Analysis reveals a weak negative correlation (Pearson R = -0.2301) between router performance and true scores, indicating that higher true scores do not necessarily translate to better routing performance. This underscores the importance of decoupled evaluation metrics.

\textbf{(3) Prompt Construction Dominance.}
$\tau=0.91$ performs best among tested thresholds; performance is relatively strong in the neighborhood around 0.91, emphasizing that prompt construction quality outweighs router performance in multi-turn dialogue systems.

\textbf{(4) Exact Partitioning is Not Necessarily Optimal.}
This can be compared from three aspects: (i) On \textbf{Evaluation Accuracy} ($\overline{si}$), when $\tau = 0.91$ with 40.89\%, it is 8.22\% higher than Conv-ID Context (32.67\%), proving that the router model trained with exact partitioning does not perform as well as \sys. (ii) On \textbf{Construction Accuracy} ($\bar{w}$), when the similarity threshold $\tau$ is 0.90, the average score of 7 individual models significantly outperforms the prompt constructed by Conv-ID Context, indicating that in multi-turn dialogue scenarios, strictly accurate context partitioning cannot be considered the optimal context segmentation method. (iii) On \textbf{Router Performance} ($P_{\text{router}}$), the average performance for both $\tau = 0.85$ and $\tau = 0.91$ surpass Conv-ID Context, meaning that the prompt constructed by Conv-ID Context also fails to train the optimal router performance.

Overall, \sys introduces limited additional computational overhead while substantially improving routing performance, demonstrating a favorable effectiveness--efficiency tradeoff for multi-turn LLM routing.

\section{Related Work}

Large language models have become increasingly heterogeneous in capability, cost, latency, and domain specialization. This heterogeneity has motivated systems that use multiple LLMs rather than committing to a single model for all inputs. Existing work can be broadly grouped into LLM ensembling and cascading, single-turn LLM routing, and context-aware routing for multi-turn dialogue.

\subsection{LLM Ensembling and Cascading}

LLM ensembling aims to improve response quality by consulting multiple models or multiple generations before producing the final answer. A simple and widely used strategy is voting or self-consistency, where multiple reasoning paths or model outputs are aggregated to reduce variance and improve reliability~\cite{li2024agents,wang2023selfconsistency}. More structured ensemble methods go beyond voting: LLM-Blender~\cite{jiang2023llm} first ranks candidate outputs with PairRanker and then synthesizes a final response with GenFuser, showing that complementary model outputs can be combined into a stronger answer.

Another line of work focuses on cascades, where models are queried sequentially according to estimated difficulty, confidence, or cost. FrugalGPT~\cite{chen2023frugalgpt} studies cost-effective model selection through cascaded calls, while language-model cascades further explore uncertainty-aware and token-level routing policies~\cite{gupta2024language,yue2024large}. Online cascade learning extends this idea to streaming or adaptive inference settings~\cite{nie2024online}. These methods reduce cost compared with invoking all available models, but they may still require multiple model calls for a single user request. In contrast, \sys follows the routing paradigm: it aims to choose one suitable candidate LLM after constructing the multi-turn prompt, thereby avoiding repeated generation from many models at inference time.

\subsection{Single-turn LLM Routing}

LLM routing selects the most suitable model for a query without necessarily calling all candidate LLMs. Early routing studies often formulate the problem as correctness prediction or reward prediction for each candidate model. Shnitzer et al.~\cite{shnitzer2023large} construct benchmark datasets for LLM routing and train model-specific binary classifiers to estimate whether each model will answer correctly. ZOOTER~\cite{lu2023routing} aligns a router with reward-model supervision and shows that learned routers can outperform static model selection. Cost-aware routing methods such as C2MAB-V~\cite{dai2024cost} further incorporate online decision making and reward signals to balance model quality and inference cost.

Recent work also studies representation-based routing. LoraRetriever~\cite{zhao2024loraretriever} routes inputs by predicting task identity and retrieving suitable LoRA modules, while Srivatsa et al.~\cite{srivatsa2024harnessing} analyze classifier-based and clustering-based routing strategies. RouterDC~\cite{chen2024RouterDC} introduces a dual contrastive learning objective that models both query-model relations and query-query relations, making it a strong single-turn router backbone. These methods provide useful supervision and representation-learning tools for model selection, and \sys builds on this direction. However, they generally assume that each routing instance is already a complete prompt. This assumption is reasonable for single-turn benchmarks, but it becomes fragile in multi-turn dialogue because the routing input itself depends on how historical context is selected and assembled.

\subsection{Context Construction for Multi-turn Dialogue}

Multi-turn dialogue introduces a context-construction problem before routing can even begin. A conversational request may depend on previous constraints, definitions, or preferences, but the full dialogue history may also contain obsolete or off-topic content. Directly concatenating all previous turns can therefore increase prompt length and introduce irrelevant information, whereas using only the current turn can omit essential context. Long-context benchmarks such as MT-Bench-101~\cite{bai2024mt} and real dialogue data such as ShareGPT~\cite{sharegpt_raw} highlight that multi-turn tasks often require tracking user intent across turns rather than treating each prompt independently.

Several adjacent research areas offer partial solutions. Retrieval-based context construction can recover long-range dependencies by searching dialogue history for semantically related turns. Sequence segmentation and similarity-based partitioning, rooted in broader sequence analysis and string-processing ideas~\cite{Aho:72,Gusfield:97,Chandra:81}, provide a way to divide a conversation into locally coherent windows. Neural encoders such as DeBERTa~\cite{he2021deberta} can then map turns into dense representations for semantic matching. Nevertheless, these techniques are usually studied as prompt-construction or retrieval components rather than as part of a routing evaluation framework. \sys integrates similarity-based window construction with router learning so that the selected model is conditioned on a prompt that is explicitly designed for multi-turn relevance.

\subsection{Evaluation of Routing Under Constructed Prompts}

Most routing evaluations focus on whether the router selects a high-scoring model for a given input. This is appropriate when the input prompt is fixed and complete, but it can be misleading when prompt construction varies. In multi-turn dialogue, a poorly constructed prompt can lower the absolute scores of all candidate LLMs even if their relative ranking remains unchanged. Conversely, a better prompt can improve all candidate responses while leaving router-selection difficulty similar. Standard aggregate metrics such as benchmark accuracy~\cite{mmlu,cmmlu,gsm8k} or reward-model score therefore conflate two factors: whether the constructed prompt preserves enough information, and whether the router chooses the best model under that prompt.

The key difference between \sys and prior routing work is this explicit separation. Instead of evaluating only the selected model's final score, \sys reports evaluation accuracy, construction accuracy, and router performance. This decoupled view makes it possible to determine whether improvements come from better prompt construction, better model selection, or both. As a result, \sys extends single-turn LLM routing to a setting where context segmentation is a first-class part of the routing problem rather than a preprocessing detail.

\section{Conclusion}

This paper presents \sys, a similarity-contractive window router that addresses two fundamental challenges in multi-turn LLM routing: information loss and information confusion caused by improper context segmentation, and attribution bias in router evaluation that conflates prompt construction quality with routing effectiveness.

To tackle these challenges, \sys introduces a similarity-based window partitioning mechanism that dynamically constructs semantically coherent prompts from dialogue history, and a decoupled evaluation framework comprising three complementary metrics: evaluation accuracy ($\overline{si}$), construction accuracy ($\bar{w}$), and router performance ($P_{\text{router}}$).
Extensive experiments on ShareGPT and MTBench demonstrate that \sys achieves \textbf{16.26\%} and \textbf{8.22\%} gains over the best individual LLM and the Conv-ID Context baseline, respectively.
On out-of-distribution tasks (PreAlgebra, MBPP, and C-EVAL), \sys further outperforms Conv-ID Context by \textbf{2.68\%} on average, demonstrating strong generalization.
Decoupled metric analysis confirms that the gains are primarily driven by improved construction accuracy, while $P_{\text{router}} > 1$ consistently validates effective model selection.
These results highlight that multi-turn LLM routing requires a joint design of context construction and evaluation, rather than a direct extension of single-turn routing methods.

Despite these advances, the current framework is evaluated on 7B/8B scale models; its behavior under larger or proprietary LLMs remains to be studied.
Future work may extend the routing framework to heterogeneous model pools spanning different scales and explore its applicability to proprietary LLMs.

\bibliographystyle{ACM-Reference-Format}
\bibliography{reference}

@book{Aho:72,
  author    = {Alfred V. Aho and Jeffrey D. Ullman},
  title     = {The Theory of Parsing, Translation, and Compiling},
  volume    = {1},
  publisher = {Prentice-Hall},
  address   = {Englewood Cliffs, NJ},
  year      = {1972}
}

@book{APA:83,
  author    = {{American Psychological Association}},
  title     = {Publications Manual},
  publisher = {American Psychological Association},
  address   = {Washington, DC},
  year      = {1983}
}

@article{Chandra:81,
  author    = {Ashok K. Chandra and Dexter Kozen and Larry J. Stockmeyer},
  title     = {Alternation},
  journal   = {Journal of the ACM},
  volume    = {28},
  number    = {1},
  pages     = {114--133},
  year      = {1981},
  doi       = {10.1145/322234.322243},
  bibsource = {dblp computer science bibliography, https://dblp.org}
}

@inproceedings{andrew2007scalable,
  author    = {Galen Andrew and Jianfeng Gao},
  title     = {Scalable Training of {L1}-Regularized Log-Linear Models},
  booktitle = {Proceedings of the 24th International Conference on Machine Learning},
  pages     = {33--40},
  publisher = {ACM},
  address   = {Corvallis, OR, USA},
  year      = {2007},
  doi       = {10.1145/1273496.1273501},
  bibsource = {dblp computer science bibliography, https://dblp.org}
}

@book{Gusfield:97,
  author    = {Dan Gusfield},
  title     = {Algorithms on Strings, Trees, and Sequences: Computer Science and Computational Biology},
  publisher = {Cambridge University Press},
  address   = {Cambridge, UK},
  year      = {1997},
  doi       = {10.1017/CBO9780511574931}
}

@article{rasooli-tetrault-2015,
  author    = {Mohammad Sadegh Rasooli and Joel R. Tetreault},
  title     = {Yara Parser: {A} Fast and Accurate Dependency Parser},
  journal   = {CoRR},
  volume    = {abs/1503.06733},
  year      = {2015},
  url       = {http://arxiv.org/abs/1503.06733},
  eprinttype = {arXiv},
  eprint    = {1503.06733},
  bibsource = {dblp computer science bibliography, https://dblp.org}
}

@article{Ando2005,
  author    = {Rie Kubota Ando and Tong Zhang},
  title     = {A Framework for Learning Predictive Structures from Multiple Tasks and Unlabeled Data},
  journal   = {Journal of Machine Learning Research},
  volume    = {6},
  pages     = {1817--1853},
  year      = {2005},
  url       = {http://jmlr.org/papers/v6/ando05a.html},
  bibsource = {dblp computer science bibliography, https://dblp.org}
}

@misc{chatgpt,
  author       = {{OpenAI}},
  title        = {ChatGPT: Optimizing Language Models for Dialogue},
  year         = {2022},
  howpublished = {\url{https://openai.com/blog/chatgpt}}
}

@article{deepseek,
  author    = {{DeepSeek-AI}},
  title     = {{DeepSeek-V3} Technical Report},
  journal   = {CoRR},
  volume    = {abs/2412.19437},
  year      = {2024},
  url       = {https://doi.org/10.48550/arXiv.2412.19437},
  doi       = {10.48550/ARXIV.2412.19437},
  eprinttype = {arXiv},
  eprint    = {2412.19437},
  bibsource = {dblp computer science bibliography, https://dblp.org}
}

@article{gemini,
  author    = {{Gemini Team}},
  title     = {Gemini: {A} Family of Highly Capable Multimodal Models},
  journal   = {CoRR},
  volume    = {abs/2312.11805},
  year      = {2023},
  url       = {https://doi.org/10.48550/arXiv.2312.11805},
  doi       = {10.48550/ARXIV.2312.11805},
  eprinttype = {arXiv},
  eprint    = {2312.11805},
  bibsource = {dblp computer science bibliography, https://dblp.org}
}

@misc{claude,
  author       = {{Anthropic}},
  title        = {Claude},
  year         = {2024},
  howpublished = {\url{https://docs.anthropic.com/en/docs/welcome}}
}

@article{li2024agents,
  author    = {Junyou Li and Qin Zhang and Yangbin Yu and Qiang Fu and Deheng Ye},
  title     = {More Agents Is All You Need},
  journal   = {CoRR},
  volume    = {abs/2402.05120},
  year      = {2024},
  url       = {https://doi.org/10.48550/arXiv.2402.05120},
  doi       = {10.48550/ARXIV.2402.05120},
  eprinttype = {arXiv},
  eprint    = {2402.05120},
  bibsource = {dblp computer science bibliography, https://dblp.org}
}

@article{deepseek-math,
  author    = {Zhihong Shao and Peiyi Wang and Qihao Zhu and Runxin Xu and Junxiao Song and Mingchuan Zhang and Y. K. Li and Y. Wu and Daya Guo},
  title     = {{DeepSeekMath}: Pushing the Limits of Mathematical Reasoning in Open Language Models},
  journal   = {CoRR},
  volume    = {abs/2402.03300},
  year      = {2024},
  url       = {https://doi.org/10.48550/arXiv.2402.03300},
  doi       = {10.48550/ARXIV.2402.03300},
  eprinttype = {arXiv},
  eprint    = {2402.03300},
  bibsource = {dblp computer science bibliography, https://dblp.org}
}

@inproceedings{wang2023selfconsistency,
  author    = {Xuezhi Wang and Jason Wei and Dale Schuurmans and Quoc V. Le and Ed H. Chi and Sharan Narang and Aakanksha Chowdhery and Denny Zhou},
  title     = {Self-Consistency Improves Chain of Thought Reasoning in Language Models},
  booktitle = {The Eleventh International Conference on Learning Representations},
  publisher = {OpenReview.net},
  address   = {Kigali, Rwanda},
  year      = {2023},
  url       = {https://openreview.net/forum?id=1PL1NIMMrw},
  bibsource = {dblp computer science bibliography, https://dblp.org}
}

@inproceedings{jiang2023llm,
  author    = {Dongfu Jiang and Xiang Ren and Bill Yuchen Lin},
  title     = {{LLM-Blender}: Ensembling Large Language Models with Pairwise Ranking and Generative Fusion},
  booktitle = {Proceedings of the 61st Annual Meeting of the Association for Computational Linguistics},
  pages     = {14165--14178},
  publisher = {Association for Computational Linguistics},
  address   = {Toronto, Canada},
  year      = {2023},
  doi       = {10.18653/V1/2023.ACL-LONG.792},
  url       = {https://aclanthology.org/2023.acl-long.792},
  bibsource = {dblp computer science bibliography, https://dblp.org}
}

@article{chen2023frugalgpt,
  author    = {Lingjiao Chen and Matei Zaharia and James Zou},
  title     = {{FrugalGPT}: How to Use Large Language Models While Reducing Cost and Improving Performance},
  journal   = {CoRR},
  volume    = {abs/2305.05176},
  year      = {2023},
  url       = {https://doi.org/10.48550/arXiv.2305.05176},
  doi       = {10.48550/ARXIV.2305.05176},
  eprinttype = {arXiv},
  eprint    = {2305.05176},
  bibsource = {dblp computer science bibliography, https://dblp.org}
}

@article{gupta2024language,
  author    = {Neha Gupta and Harikrishna Narasimhan and Wittawat Jitkrittum and Ankit Singh Rawat and Aditya Krishna Menon and Sanjiv Kumar},
  title     = {Language Model Cascades: Token-Level Uncertainty and Beyond},
  journal   = {CoRR},
  volume    = {abs/2404.10136},
  year      = {2024},
  url       = {https://doi.org/10.48550/arXiv.2404.10136},
  doi       = {10.48550/ARXIV.2404.10136},
  eprinttype = {arXiv},
  eprint    = {2404.10136},
  bibsource = {dblp computer science bibliography, https://dblp.org}
}

@inproceedings{nie2024online,
  author    = {Lunyiu Nie and Zhimin Ding and Erdong Hu and Christopher M. Jermaine and Swarat Chaudhuri},
  title     = {Online Cascade Learning for Efficient Inference over Streams},
  booktitle = {Proceedings of the 41st International Conference on Machine Learning},
  series    = {Proceedings of Machine Learning Research},
  volume    = {235},
  pages     = {38071--38090},
  publisher = {PMLR},
  address   = {Vienna, Austria},
  year      = {2024},
  url       = {https://proceedings.mlr.press/v235/nie24a.html},
  bibsource = {dblp computer science bibliography, https://dblp.org}
}

@article{shnitzer2023large,
  author    = {Tal Shnitzer and Anthony Ou and M{\'{\i}}rian Silva and Kate Soule and Yuekai Sun and Justin Solomon and Neil Thompson and Mikhail Yurochkin},
  title     = {Large Language Model Routing with Benchmark Datasets},
  journal   = {CoRR},
  volume    = {abs/2309.15789},
  year      = {2023},
  url       = {https://doi.org/10.48550/arXiv.2309.15789},
  doi       = {10.48550/ARXIV.2309.15789},
  eprinttype = {arXiv},
  eprint    = {2309.15789},
  bibsource = {dblp computer science bibliography, https://dblp.org}
}

@inproceedings{lu2023routing,
  author    = {Keming Lu and Hongyi Yuan and Runji Lin and Junyang Lin and Zheng Yuan and Chang Zhou and Jingren Zhou},
  title     = {Routing to the Expert: Efficient Reward-Guided Ensemble of Large Language Models},
  booktitle = {Proceedings of the 2024 Conference of the North American Chapter of the Association for Computational Linguistics: Human Language Technologies},
  pages     = {1964--1974},
  publisher = {Association for Computational Linguistics},
  address   = {Mexico City, Mexico},
  year      = {2024},
  doi       = {10.18653/V1/2024.NAACL-LONG.109},
  url       = {https://aclanthology.org/2024.naacl-long.109},
  bibsource = {dblp computer science bibliography, https://dblp.org}
}

@inproceedings{yue2024large,
  author    = {Murong Yue and Jie Zhao and Min Zhang and Liang Du and Ziyu Yao},
  title     = {Large Language Model Cascades with Mixture of Thought Representations for Cost-Efficient Reasoning},
  booktitle = {The Twelfth International Conference on Learning Representations},
  publisher = {OpenReview.net},
  address   = {Vienna, Austria},
  year      = {2024},
  url       = {https://openreview.net/forum?id=6okaSfANzh},
  bibsource = {dblp computer science bibliography, https://dblp.org}
}

@inproceedings{zhao2024loraretriever,
  author    = {Ziyu Zhao and Leilei Gan and Guoyin Wang and Wangchunshu Zhou and Hongxia Yang and Kun Kuang and Fei Wu},
  title     = {{LoraRetriever}: Input-Aware {LoRA} Retrieval and Composition for Mixed Tasks in the Wild},
  booktitle = {Findings of the Association for Computational Linguistics: ACL 2024},
  pages     = {4447--4462},
  publisher = {Association for Computational Linguistics},
  address   = {Bangkok, Thailand},
  year      = {2024},
  doi       = {10.18653/V1/2024.FINDINGS-ACL.263},
  url       = {https://aclanthology.org/2024.findings-acl.263},
  bibsource = {dblp computer science bibliography, https://dblp.org}
}

@article{srivatsa2024harnessing,
  author    = {{KV} Aditya Srivatsa and Kaushal Kumar Maurya and Ekaterina Kochmar},
  title     = {Harnessing the Power of Multiple Minds: Lessons Learned from {LLM} Routing},
  journal   = {CoRR},
  volume    = {abs/2405.00467},
  year      = {2024},
  url       = {https://doi.org/10.48550/arXiv.2405.00467},
  doi       = {10.48550/ARXIV.2405.00467},
  eprinttype = {arXiv},
  eprint    = {2405.00467},
  bibsource = {dblp computer science bibliography, https://dblp.org}
}

@inproceedings{chen2024RouterDC,
  author    = {Shuhao Chen and Weisen Jiang and Baijiong Lin and James T. Kwok and Yu Zhang},
  title     = {{RouterDC}: Query-Based Router by Dual Contrastive Learning for Assembling Large Language Models},
  booktitle = {Advances in Neural Information Processing Systems},
  volume    = {37},
  pages     = {66305--66328},
  publisher = {Curran Associates, Inc.},
  address   = {Vancouver, BC, Canada},
  year      = {2024},
  url       = {https://proceedings.neurips.cc/paper_files/paper/2024/hash/7a641b8ec86162fc875fb9f6456a542f-Abstract-Conference.html},
  bibsource = {dblp computer science bibliography, https://dblp.org}
}

@article{dai2024cost,
  author    = {Xiangxiang Dai and Jin Li and Xutong Liu and Anqi Yu and John C. S. Lui},
  title     = {Cost-Effective Online Multi-{LLM} Selection with Versatile Reward Models},
  journal   = {CoRR},
  volume    = {abs/2405.16587},
  year      = {2024},
  url       = {https://doi.org/10.48550/arXiv.2405.16587},
  doi       = {10.48550/ARXIV.2405.16587},
  eprinttype = {arXiv},
  eprint    = {2405.16587},
  bibsource = {dblp computer science bibliography, https://dblp.org}
}

@article{touvron2023llama,
  author    = {Hugo Touvron and Thibaut Lavril and Gautier Izacard and Xavier Martinet and Marie{-}Anne Lachaux and Timoth{\'{e}}e Lacroix and Baptiste Rozi{\`{e}}re and Naman Goyal and Eric Hambro and Faisal Azhar and Aur{\'{e}}lien Rodriguez and Armand Joulin and Edouard Grave and Guillaume Lample},
  title     = {{LLaMA}: Open and Efficient Foundation Language Models},
  journal   = {CoRR},
  volume    = {abs/2302.13971},
  year      = {2023},
  url       = {https://doi.org/10.48550/arXiv.2302.13971},
  doi       = {10.48550/ARXIV.2302.13971},
  eprinttype = {arXiv},
  eprint    = {2302.13971},
  bibsource = {dblp computer science bibliography, https://dblp.org}
}

@inproceedings{cmmlu,
  author    = {Haonan Li and Yixuan Zhang and Fajri Koto and Yifei Yang and Hai Zhao and Yeyun Gong and Nan Duan and Timothy Baldwin},
  title     = {{CMMLU}: Measuring Massive Multitask Language Understanding in Chinese},
  booktitle = {Findings of the Association for Computational Linguistics: ACL 2024},
  pages     = {11260--11285},
  publisher = {Association for Computational Linguistics},
  address   = {Bangkok, Thailand},
  year      = {2024},
  doi       = {10.18653/V1/2024.FINDINGS-ACL.671},
  url       = {https://aclanthology.org/2024.findings-acl.671},
  bibsource = {dblp computer science bibliography, https://dblp.org}
}

@article{gsm8k,
  author    = {Karl Cobbe and Vineet Kosaraju and Mohammad Bavarian and Mark Chen and Heewoo Jun and Lukasz Kaiser and Matthias Plappert and Jerry Tworek and Jacob Hilton and Reiichiro Nakano and Christopher Hesse and John Schulman},
  title     = {Training Verifiers to Solve Math Word Problems},
  journal   = {CoRR},
  volume    = {abs/2110.14168},
  year      = {2021},
  url       = {https://arxiv.org/abs/2110.14168},
  eprinttype = {arXiv},
  eprint    = {2110.14168},
  bibsource = {dblp computer science bibliography, https://dblp.org}
}

@inproceedings{mmlu,
  author    = {Dan Hendrycks and Collin Burns and Steven Basart and Andy Zou and Mantas Mazeika and Dawn Song and Jacob Steinhardt},
  title     = {Measuring Massive Multitask Language Understanding},
  booktitle = {The Ninth International Conference on Learning Representations},
  publisher = {OpenReview.net},
  address   = {Virtual Event, Austria},
  year      = {2021},
  url       = {https://openreview.net/forum?id=d7KBjmI3GmQ},
  bibsource = {dblp computer science bibliography, https://dblp.org}
}

@article{jiang2023mistral,
  author    = {Albert Q. Jiang and Alexandre Sablayrolles and Arthur Mensch and Chris Bamford and Devendra Singh Chaplot and Diego de Las Casas and Florian Bressand and Gianna Lengyel and Guillaume Lample and Lucile Saulnier and L{\'{e}}lio Renard Lavaud and Marie{-}Anne Lachaux and Pierre Stock and Teven Le Scao and Thibaut Lavril and Thomas Wang and Timoth{\'{e}}e Lacroix and William El Sayed},
  title     = {Mistral 7B},
  journal   = {CoRR},
  volume    = {abs/2310.06825},
  year      = {2023},
  url       = {https://doi.org/10.48550/arXiv.2310.06825},
  doi       = {10.48550/ARXIV.2310.06825},
  eprinttype = {arXiv},
  eprint    = {2310.06825},
  bibsource = {dblp computer science bibliography, https://dblp.org}
}

@inproceedings{yu2023metamath,
  author    = {Longhui Yu and Weisen Jiang and Han Shi and Jincheng Yu and Zhengying Liu and Yu Zhang and James T. Kwok and Zhenguo Li and Adrian Weller and Weiyang Liu},
  title     = {{MetaMath}: Bootstrap Your Own Mathematical Questions for Large Language Models},
  booktitle = {The Twelfth International Conference on Learning Representations},
  publisher = {OpenReview.net},
  address   = {Vienna, Austria},
  year      = {2024},
  url       = {https://openreview.net/forum?id=N8N0hgNDRt},
  bibsource = {dblp computer science bibliography, https://dblp.org}
}

@article{tunstall2023zephyr,
  author    = {Lewis Tunstall and Edward Beeching and Nathan Lambert and Nazneen Rajani and Kashif Rasul and Younes Belkada and Shengyi Huang and Leandro von Werra and Cl{\'{e}}mentine Fourrier and Nathan Habib and Nathan Sarrazin and Omar Sanseviero and Alexander M. Rush and Thomas Wolf},
  title     = {Zephyr: Direct Distillation of {LM} Alignment},
  journal   = {CoRR},
  volume    = {abs/2310.16944},
  year      = {2023},
  url       = {https://doi.org/10.48550/arXiv.2310.16944},
  doi       = {10.48550/ARXIV.2310.16944},
  eprinttype = {arXiv},
  eprint    = {2310.16944},
  bibsource = {dblp computer science bibliography, https://dblp.org}
}

@inproceedings{rafailov2024direct,
  author    = {Rafael Rafailov and Archit Sharma and Eric Mitchell and Christopher D. Manning and Stefano Ermon and Chelsea Finn},
  title     = {Direct Preference Optimization: Your Language Model is Secretly a Reward Model},
  booktitle = {Advances in Neural Information Processing Systems},
  volume    = {36},
  pages     = {53728--53741},
  publisher = {Curran Associates, Inc.},
  address   = {New Orleans, LA, USA},
  year      = {2023},
  url       = {https://proceedings.neurips.cc/paper_files/paper/2023/hash/a85b405ed65c6477a4fe8302b5e06ce7-Abstract-Conference.html},
  bibsource = {dblp computer science bibliography, https://dblp.org}
}

@misc{Chinese-Mistral,
  author       = {Chen Zhou and Yuqi Bai},
  title        = {{Chinese-Mistral}: An Efficient and Effective Chinese Large Language Model},
  year         = {2024},
  howpublished = {\url{https://github.com/THU-ESIS/Chinese-Mistral}}
}

@misc{dolphin-2.6-mistral-7b,
  author       = {{Cognitive Computations}},
  title        = {cognitivecomputations/dolphin-2.6-mistral-7b},
  year         = {2024},
  howpublished = {\url{https://huggingface.co/cognitivecomputations/dolphin-2.6-mistral-7b}}
}

@article{dubey2024llama,
  author    = {{Llama Team}},
  title     = {The Llama 3 Herd of Models},
  journal   = {CoRR},
  volume    = {abs/2407.21783},
  year      = {2024},
  url       = {https://doi.org/10.48550/arXiv.2407.21783},
  doi       = {10.48550/ARXIV.2407.21783},
  eprinttype = {arXiv},
  eprint    = {2407.21783},
  bibsource = {dblp computer science bibliography, https://dblp.org}
}

@misc{dolphin-2.9-llama3-8b,
  author       = {{Cognitive Computations}},
  title        = {cognitivecomputations/dolphin-2.9-llama3-8b},
  year         = {2024},
  howpublished = {\url{https://huggingface.co/cognitivecomputations/dolphin-2.9-llama3-8b}}
}

@inproceedings{bai2024mt,
  author    = {Ge Bai and Jie Liu and Xingyuan Bu and Yancheng He and Jiaheng Liu and Zhanhui Zhou and Zhuoran Lin and Wenbo Su and Tiezheng Ge and Bo Zheng and Wanli Ouyang},
  title     = {{MT}-Bench-101: {A} Fine-Grained Benchmark for Evaluating Large Language Models in Multi-Turn Dialogues},
  booktitle = {Proceedings of the 62nd Annual Meeting of the Association for Computational Linguistics},
  pages     = {7421--7454},
  publisher = {Association for Computational Linguistics},
  address   = {Bangkok, Thailand},
  year      = {2024},
  doi       = {10.18653/V1/2024.ACL-LONG.401},
  url       = {https://aclanthology.org/2024.acl-long.401}
}

@misc{sharegpt_raw,
  author       = {Philipp Schmid},
  title        = {ShareGPT 90k Raw Dataset},
  year         = {2023},
  howpublished = {\url{https://huggingface.co/datasets/philschmid/sharegpt-raw/tree/main/sharegpt_90k_raw_dataset}}
}

@misc{eval-harness,
  author       = {Leo Gao and Jonathan Tow and Baber Abbasi and Stella Biderman and Sid Black and Anthony DiPofi and Charles Foster and Laurence Golding and Jeffrey Hsu and Alain Le Noac'h and Haonan Li and Kyle McDonell and Niklas Muennighoff and Chris Ociepa and Jason Phang and Laria Reynolds and Hailey Schoelkopf and Aviya Skowron and Lintang Sutawika and Eric Tang and Anish Thite and Ben Wang and Kevin Wang and Andy Zou},
  title        = {A Framework for Few-Shot Language Model Evaluation},
  year         = {2023},
  doi          = {10.5281/zenodo.10256836},
  howpublished = {\url{https://zenodo.org/records/10256836}}
}

@inproceedings{he2021deberta,
  author    = {Pengcheng He and Xiaodong Liu and Jianfeng Gao and Weizhu Chen},
  title     = {{DeBERTa}: Decoding-Enhanced {BERT} with Disentangled Attention},
  booktitle = {The Ninth International Conference on Learning Representations},
  publisher = {OpenReview.net},
  address   = {Virtual Event, Austria},
  year      = {2021},
  url       = {https://openreview.net/forum?id=XPZIaotutsD},
  bibsource = {dblp computer science bibliography, https://dblp.org}
}

@inproceedings{loshchilov2018decoupled,
  author    = {Ilya Loshchilov and Frank Hutter},
  title     = {Decoupled Weight Decay Regularization},
  booktitle = {International Conference on Learning Representations},
  publisher = {OpenReview.net},
  address   = {New Orleans, LA, USA},
  year      = {2019},
  url       = {https://openreview.net/forum?id=Bkg6RiCqY7},
  bibsource = {dblp computer science bibliography, https://dblp.org}
}

@article{kullback1951information,
  author  = {Solomon Kullback and Richard A. Leibler},
  title   = {On Information and Sufficiency},
  journal = {The Annals of Mathematical Statistics},
  volume  = {22},
  number  = {1},
  pages   = {79--86},
  year    = {1951},
  doi     = {10.1214/aoms/1177729694}
}

@article{touvron2023llama2,
  author    = {Hugo Touvron and Louis Martin and Kevin Stone and Peter Albert and Amjad Almahairi and Yasmine Babaei and Nikolay Bashlykov and Soumya Batra and Prajjwal Bhargava and Shruti Bhosale and Dan Bikel and Lukas Blecher and Cristian Canton{-}Ferrer and Moya Chen and Guillem Cucurull and David Esiobu and Jude Fernandes and Jeremy Fu and Wenyin Fu and Brian Fuller and Cynthia Gao and Vedanuj Goswami and Naman Goyal and Anthony Hartshorn and Saghar Hosseini and Rui Hou and Hakan Inan and Marcin Kardas and Viktor Kerkez and Madian Khabsa and Isabel Kloumann and Artem Korenev and Punit Singh Koura and Marie{-}Anne Lachaux and Thibaut Lavril and Jenya Lee and Diana Liskovich and Yinghai Lu and Yuning Mao and Xavier Martinet and Todor Mihaylov and Pushkar Mishra and Igor Molybog and Yixin Nie and Andrew Poulton and Jeremy Reizenstein and Rashi Rungta and Kalyan Saladi and Alan Schelten and Ruan Silva and Eric Michael Smith and Ranjan Subramanian and Xiaoqing Ellen Tan and Binh Tang and Ross Taylor and Adina Williams and Jian Xiang Kuan and Puxin Xu and Zheng Yan and Iliyan Zarov and Yuchen Zhang and Angela Fan and Melanie Kambadur and Sharan Narang and Aur{\'{e}}lien Rodriguez and Robert Stojnic and Sergey Edunov and Thomas Scialom},
  title     = {Llama 2: Open Foundation and Fine-Tuned Chat Models},
  journal   = {CoRR},
  volume    = {abs/2307.09288},
  year      = {2023},
  url       = {https://doi.org/10.48550/arXiv.2307.09288},
  doi       = {10.48550/ARXIV.2307.09288},
  eprinttype = {arXiv},
  eprint    = {2307.09288},
  bibsource = {dblp computer science bibliography, https://dblp.org}
}

@article{li2025aisearch,
  title={Towards ai search paradigm},
  author={Li, Yuchen and Cai, Hengyi and Kong, Rui and Chen, Xinran and Chen, Jiamin and Yang, Jun and Zhang, Haojie and Li, Jiayi and Wu, Jiayi and Chen, Yiqun and others},
  journal={arXiv preprint arXiv:2506.17188},
  year={2025}
}

@inproceedings{li2026retain,
  title={Retain to Refine: Adaptive Online Question Answering via Query Routing and Long-Short Memory},
  author={Li, Yuchen and Chen, Jiamin and Chen, Xinran and Li, Zhiyu and Zhang, Haojie and Kong, Rui and Li, Jiayi and Ma, Xinyu and Cai, Hengyi and Su, Lixin and others},
  booktitle={Proceedings of the 32nd ACM SIGKDD Conference on Knowledge Discovery and Data Mining V. 1},
  pages={2312--2322},
  year={2026}
}

@inproceedings{li2026probe,
  title={Probe-and-fetch: Dynamic KV cache pruning for accelerated long-context inference in web-scale AI search},
  author={Li, Yuchen and Kong, Rui and Chen, Xinran and Zhang, Chengzhe and Chen, Jiamin and Deng, Cheng and Ma, Xinyu and Zhang, Haojie and Peng, Tianhao and Cai, Hengyi and others},
  booktitle={Proceedings of the ACM Web Conference 2026},
  pages={8127--8137},
  year={2026}
}

@inproceedings{chen2025multi,
  title={Multi-agent proactive information seeking with adaptive llm orchestration for non-factoid question answering},
  author={Chen, Xinran and Li, Yuchen and Cai, Hengyi and Ma, Zhuoran and Chen, Xuanang and Xiong, Haoyi and Wang, Shuaiqiang and He, Ben and Sun, Le and Yin, Dawei},
  booktitle={Proceedings of the 31st ACM SIGKDD Conference on Knowledge Discovery and Data Mining V. 2},
  pages={4341--4352},
  year={2025}
}

@article{li2026flexspec,
  title={Flexspec: Frozen drafts meet evolving targets in edge-cloud collaborative llm speculative decoding},
  author={Li, Yuchen and Kong, Rui and Lyu, Zhonghao and Li, Qiyang and Chen, Xinran and Cai, Hengyi and Yan, Lingyong and Wang, Shuaiqiang and Zhao, Jiashu and Zhu, Guangxu and others},
  journal={IEEE Transactions on Mobile Computing},
  year={2026},
  publisher={IEEE}
}

\end{document}